\documentclass{article} 
\usepackage{iclr2027_conference,times}

\usepackage{amsmath,amsfonts,bm}

\def\eqref#1{equation~\ref{#1}}

\def\1{\bm{1}}

\DeclareMathAlphabet{\mathsfit}{\encodingdefault}{\sfdefault}{m}{sl}
\SetMathAlphabet{\mathsfit}{bold}{\encodingdefault}{\sfdefault}{bx}{n}

\usepackage{hyperref}
\usepackage{url}

\usepackage{booktabs}
\usepackage{tabularx}
\usepackage{array}
\usepackage{multirow}
\usepackage[table]{xcolor}
\usepackage{threeparttable}
\usepackage{colortbl}
\usepackage{adjustbox}
\usepackage{makecell}
\usepackage{subcaption}
\usepackage{wrapfig}

\newlength{\omnigaiatablewidth}
\definecolor{textgroup}{RGB}{240,246,252}
\definecolor{avgroup}{RGB}{252,245,237}
\definecolor{groupbar}{RGB}{238,239,241}
\definecolor{deltapos}{RGB}{36,120,86}
\definecolor{deltaneg}{RGB}{176,62,62}

\newcommand{\posd}[1]{%
  {\tiny\bfseries\boldmath\color{deltapos}$(+#1)$}%
}

\newcommand{\negd}[1]{%
  {\tiny\bfseries\boldmath\color{deltaneg}$(-#1)$}%
}

\newcommand{\zerod}{%
  {\tiny\bfseries\boldmath\color{black!55}$(+0.00)$}%
}

\title{Text-Centric Post-Training for Omni-Modal Reasoning}

\author{Ziyang Cheng\thanks{Equal contribution.}, Yuhao Wang\footnotemark[1], Hongcheng Liu\footnotemark[1], Qimin Wu, \\
\textbf{Jingru Fan, Chen Qian, Yanfeng Wang \& Yu Wang\thanks{Corresponding author.}} \\
Shanghai Jiao Tong University \\
\texttt{\{muye12,colane,hongcheng\_liu,wuqimin1260\}@sjtu.edu.cn} \\
\texttt{\{fanjingru510,qianc,wangyanfeng622,yuwangsjtu\}@sjtu.edu.cn}
}

\iclrfinalcopy 
\begin{document}

\maketitle
\pagestyle{plain}

\begin{abstract}
Improving joint audio-visual reasoning in Omni Large Language Models typically incurs substantial data construction and training costs.
Our diagnostics reveal multi-hop reasoning difficulties despite correct answers to all corresponding single-hop questions and suggest partial decoupling in the local optimization of perception and reasoning objectives.
This motivates post-training with different emphases on these capabilities.
Text-only reasoning training yields gains across data sources, model scales, and families.
With the best-performing text-only configuration, supervised fine-tuning followed by reinforcement learning (RL) raises Qwen2.5-Omni-7B's geometric mean of nine reasoning scores by 25.83\% over the base model, outperforming the complete native audio-visual route with 56.6\% fewer GPU-hours.
Training on data synthesized entirely by a text-only LLM raises this geometric mean by 21.01\% without audio-visual data in construction or training.
However, text-only training degrades perception.
We therefore propose a text-centric post-training paradigm: text-only training provides the main reasoning optimization, and reduced-data native audio-visual RL then refines perception.
Refinement uses about 90\% fewer input tokens than full-data audio-visual RL, restores perception above the base level, and retains 93.5\% of the best-performing text-only pipeline's reasoning gain.
\end{abstract}

\section{Introduction}
Omni Large Language Models (Omni-LLMs) have made remarkable progress in unified multimodal perception and understanding, enabling a single model to process text, audio, images, and video \citep{Qwen2.5-Omni,xu2025qwen3,team2026qwen3,deshmukh2026nemotron,team2026gemma}. However, strong perceptual capabilities do not necessarily translate into reliable multimodal reasoning \citep{yao2026lens,wang2026compose}. Existing Omni-LLMs still struggle with complex audio-visual tasks that require combining information across modalities and over time \citep{chen2025omnixr,dai2026latentomni}.

\suppressfloats[t]
\begin{figure}[t]
    \centering
    \includegraphics[width=0.94\linewidth]{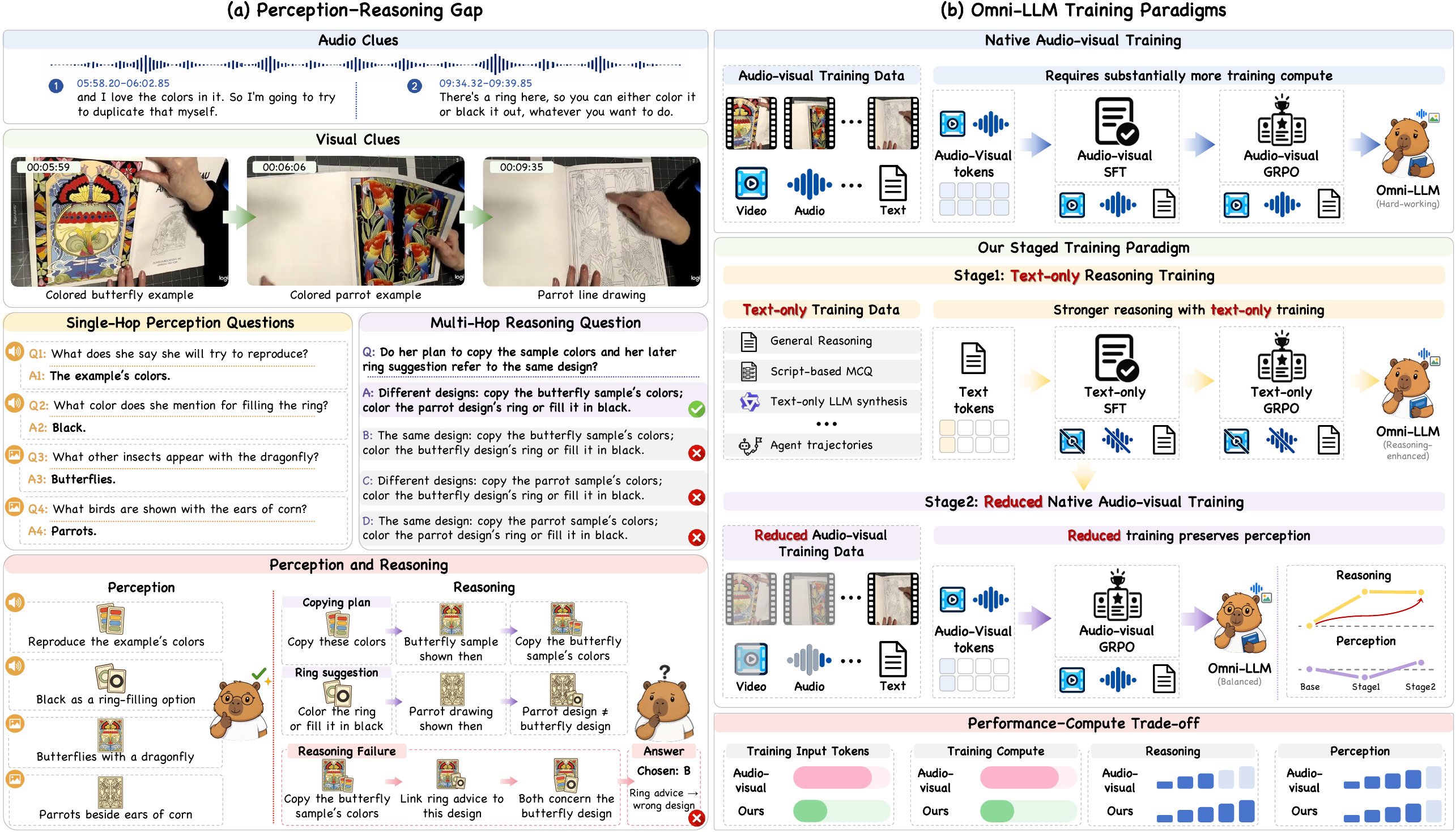}
    \caption{
    (a) A diagnostic case of multi-hop reasoning failure despite correct answers to corresponding single-hop questions.
    (b) Text-only training provides the main reasoning optimization, followed by reduced-data native audio-visual RL to restore perception while retaining most reasoning gains.
    }
    \label{fig:intro}
\end{figure}

To address this limitation, existing approaches typically use specially constructed multimodal reasoning data for supervised fine-tuning (SFT) or reinforcement learning (RL) \citep{xing2025echoink,yang2025humanomniv2,kulkarni2026avatar,araujo2026avrt}. Constructing such data requires annotations consistent with audio-visual content and adequate coverage of infrequent categories in long-tailed distributions \citep{cai2026omnivideo,chen2026omnivideo}. These requirements and multimodal training costs limit scalability \citep{guan2026thinkomni}. Meanwhile, text-only reasoning training has improved performance on separate visual and audio tasks \citep{liu2025x,bousselham2026vold,rouditchenko2025omni}. However, transfer to these tasks does not establish its effectiveness for joint audio-visual reasoning, which requires integrating complementary evidence across modalities and over time. This motivates us to ask: Can Omni-LLM post-training be organized to target perception and reasoning separately while reducing reliance on native audio-visual training?

In Section~\ref{sec:preliminary}, we investigate this possibility through a diagnostic study. Across nine Omni-LLMs, multi-hop reasoning remains difficult despite correct answers to all corresponding single-hop perception questions. A diagnostic case is shown in Figure~\ref{fig:intro}(a). Beyond this performance gap, we examine whether perception and reasoning admit different optimization directions. Projected gradient updates retain most of the target loss reduction while substantially reducing changes to the other objective. These responses support partial decoupling in the local optimization of the two mean-loss objectives, motivating training strategies with different emphases on perception and reasoning.

Building on these findings, we explore organizing Omni-LLM training around the capabilities to be improved, rather than downstream input modalities. We examine text-only reasoning training across data sources. On Qwen2.5-Omni-7B, SFT followed by RL using the best-performing text source improves all nine reasoning scores, raising their average by 9.59 points and geometric mean by 25.83\% over the base model. It outperforms the complete native audio-visual route with 56.6\% fewer GPU-hours. Gains extend across the evaluated Qwen2.5-Omni, Qwen3-Omni \citep{xu2025qwen3}, and Gemma 4 \citep{team2026gemma} models of different scales. Crucially, a text-only LLM generates fictional scenario descriptions and reasoning questions and answers for SFT and RL, without native audio-visual data in construction or training. Training on these fully synthetic data raises the average reasoning score by 7.70 points and geometric mean by 21.01\% over the same base model.

However, text-only reasoning training degrades perception. We therefore propose the text-centric post-training paradigm in Figure~\ref{fig:intro}(b): text-only SFT followed by RL provides the main reasoning optimization, and reduced-data native audio-visual RL then refines perception. On Qwen2.5-Omni-7B, refinement uses about 90\% fewer input tokens than full-data audio-visual RL. It restores perception above the base level while retaining 93.5\% of the best-performing text-only pipeline's gain in the geometric mean of reasoning scores, addressing the performance--compute trade-off.

Our contributions are as follows:
\begin{itemize}
    \item We conduct a diagnostic study revealing a gap between single-hop perception and multi-hop reasoning and suggesting partial decoupling in their local optimization.
    \item We systematically evaluate text-only reasoning training across data sources, model scales, and families, demonstrating omni-modal reasoning gains even with fully synthetic text requiring no audio-visual data in construction or training. Text-only training outperforms the complete native audio-visual route with lower training compute.
    \item We propose and validate a text-centric post-training paradigm: text-only training provides the main reasoning optimization, and reduced-data native audio-visual RL then restores perception while retaining most of the reasoning gains.
\end{itemize}

\section{Related Work}
\label{sec:related_work}

\paragraph{Text-only reasoning transfer.}
We study joint audio-visual reasoning across modalities and time, whereas text-only transfer studies address visual \citep{li2025temporal,liu2025x,bousselham2026vold} or audio tasks \citep{rouditchenko2025omni}. Our text-only post-training strengthens reasoning within Omni-LLMs without external inference-time reasoners \citep{zhang2025seeingeye}. Unlike text-derived multimodal supervision \citep{yu2025unicorn,cheng2026omni}, our fully synthetic setting uses textual SFT and RL data generated by a text-only LLM without native audio-visual inputs.

\paragraph{Perception and reasoning in multimodal models.}
Cross-modal evaluations reveal perception and integration limitations \citep{wang2026xmodbench,wang2026avi,mukherjee2026c}. Methods use multimodal training \citep{liu-etal-2026-cross,jia2026omnijigsaw,liu2026boosting,liu2026anchornet} or active inspection \citep{xu2026agentic,chen2026omnireasoner}. VLM post-training balances perception and reasoning \citep{peng2026lmm,chen2026revisual,wu2026seeing}. Our diagnostics motivate text-only post-training that improves joint audio-visual reasoning without native audio-visual training data.

\section{Preliminary Study}
\label{sec:preliminary}
For this diagnostic study, we construct questions from annotations of real audio-visual data and evaluate nine Omni-LLMs on the corresponding inputs. Single-hop perception questions assess the direct extraction of information. From the common subset answered correctly by all nine models, we construct 2-hop and 3-hop reasoning questions that combine the corresponding facts across audio and vision in two and three steps, respectively (Figure~\ref{fig:intro}(a)). Details are provided in Appendix~\ref{sec:appendix_diagnostic_questions}.

\begin{figure}[htbp]
    \centering
    \input{figures/preliminary_overview_dimensions.tex}
    \captionsetup{justification=centering,singlelinecheck=false}
    \captionsetup[subfigure]{justification=centering,singlelinecheck=false}
    \begin{minipage}{0.8\linewidth}
    \includegraphics[width=\linewidth]{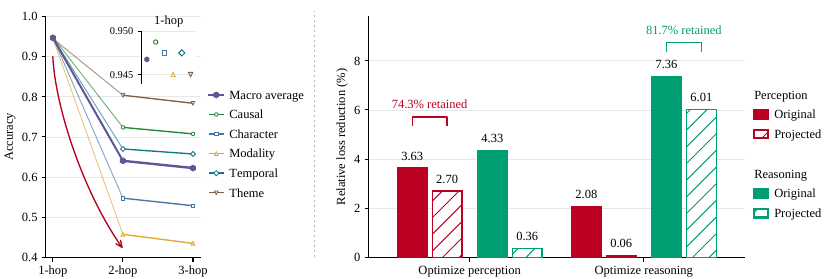}

    \makebox[\linewidth][l]{%
        \hspace*{\FigureTwoCaptionACenter\linewidth}%
        \makebox[0pt][c]{%
            \begin{subfigure}[t]{0.38\linewidth}
                \centering
                \caption{Accuracy across hops.}
                \label{fig:hop_gap}
            \end{subfigure}%
        }%
        \hspace*{\FigureTwoCaptionSeparation\linewidth}%
        \makebox[0pt][c]{%
            \begin{subfigure}[t]{0.55\linewidth}
                \centering
                \caption{Gradient projection.}
                \label{fig:local_optimization}
            \end{subfigure}%
        }%
    }
    \end{minipage}
    \caption{Perception--reasoning gap and local update responses.}
    \label{fig:preliminary}
\end{figure}

\subsection{The Gap Between Perception and Reasoning}

Figure~\ref{fig:hop_gap} reports macro-averaged accuracy across models, with 94.7\% on the original 1-hop set. Despite correct answers to every corresponding single-hop question, accuracy on the resulting multi-hop questions is only 64.1\% for 2-hop and 62.3\% for 3-hop. Across all five domains, these results reveal a gap between single-hop perception and multi-hop reasoning: recognizing individual facts does not ensure that models can combine them to reach the correct answer.

\subsection{Partial Decoupling in Perception and Reasoning Optimization}
\label{sec:preliminary_gradients}

Beyond the performance gap, we measure improvement on one objective while constraining changes to the other. In Qwen2.5-Omni-3B \citep{Qwen2.5-Omni}, we define $L_E$ and $L_R$ as the mean cross-entropy losses for predicting the correct answer token on single-hop perception questions and 2-hop and 3-hop reasoning questions, respectively. Under five-fold cross-validation by video, each split defines these objectives on its four training folds. We compute gradients for the Thinker's Transformer decoder layers and final normalization layer, with all other parameters fixed.

Motivated by orthogonal gradient projection \citep{huang2025digrap,qiu2026splitlora}, we compare original gradient updates with updates constrained to leave the other training-fold mean loss unchanged to first order. At the initial checkpoint $\theta_0$, let $g_t=\nabla L_t(\theta_0)$ for distinct $t,s\in\{E,R\}$ and define
\begin{equation}
\label{eq:local_projection}
r_{t\perp s}=g_t-\frac{g_s^\top g_t}{\|g_s\|_2^2}g_s,
\qquad
\theta'_{t\perp s}=\theta_0-\eta\frac{r_{t\perp s}}{\|r_{t\perp s}\|_2}.
\end{equation}
The original update is $\theta'_t=\theta_0-\eta g_t/\|g_t\|_2$. Both updates start from $\theta_0$, with parameter changes of the same Euclidean norm $\eta$. We pool held-out loss reductions across folds and express them relative to each objective's pooled initial loss; positive values indicate decreases and negative values indicate increases. Implementation details and supplementary analyses appear in Appendix~\ref{sec:appendix_gradient_analysis}.

In Figure~\ref{fig:local_optimization}, the projected reasoning update reduces reasoning loss by 6.01\% (81.7\% of the original reduction) and perception loss by 0.06\% (original update: 2.08\%). The projected perception update retains 74.3\% of the original target loss reduction and reduces reasoning loss by 0.36\% (original update: 4.33\%). These responses support partial decoupling of the two mean-loss objectives in local optimization, motivating different post-training emphases on perception and reasoning.

\section{Study Design}
\label{sec:study_design}

Building on the findings in Section~\ref{sec:preliminary}, we systematically address five research questions (RQs) to develop a post-training paradigm that improves omni-modal reasoning with lower training compute:

\begin{itemize}
    \item \textbf{(RQ1) Reasoning Gains and Training Compute.}
    To what extent does text-only reasoning training improve omni-modal reasoning performance across diverse tasks and models, and how much training compute does it require?

    \item \textbf{(RQ2) Choice of Training Data.}
    How does the choice of training data affect the reasoning gains from text-only reasoning training on omni-modal tasks?

    \item \textbf{(RQ3) Synthetic Training Data.}
    Can text-only reasoning training still yield omni-modal reasoning gains when all training data are synthesized by a text-only LLM?

    \item \textbf{(RQ4) Perceptual Effects.}
    How does text-only reasoning training affect the perceptual capabilities of Omni-LLMs relative to native audio-visual training?

    \item \textbf{(RQ5) Training Paradigm.}
    How should text-centric post-training be staged to improve omni-modal reasoning while maintaining perceptual capabilities?

\end{itemize}

\subsection{Training Setup}
\label{sec:training_setup}

\paragraph{Models.}
We conduct our main experiments on Qwen2.5-Omni-7B \citep{Qwen2.5-Omni}. To examine gains across model scales and families, we apply the same text-only training pipeline to Qwen2.5-Omni-3B \citep{Qwen2.5-Omni}, Gemma-4-E4B-IT, and Gemma-4-E2B-IT \citep{team2026gemma}. We also evaluate text-only SFT on Qwen3-Omni-30B-A3B-Instruct \citep{xu2025qwen3}.

\paragraph{Training details.}
Text-only and native audio-visual post-training start from the same base model and follow SFT with Group Relative Policy Optimization (GRPO) \citep{shao2024deepseekmath}. We adopt the best-performing text-only configuration from the RQ2 comparison as the default, using Script-based MCQs in both stages and CoT supervision for SFT. Table~\ref{tab:training_data} lists data types and sources for each RQ.
An overview of the training setup is provided in Appendix~\ref{sec:appendix_training_setup}.

\subsection{Evaluation Setup}
\label{sec:evaluation_setup}

\paragraph{Benchmarks.} 
For RQ1--RQ3, we evaluate omni-modal reasoning on Daily-Omni \citep{zhou2025daily}, Omni-Cloze \citep{ma2026omni}, OmniVideoBench \citep{li2026omnivideobench}, SocialOmni \citep{xie2026socialomni}, Video-Holmes \citep{cheng2025video}, Video-MME-v2 \citep{fu2026video}, and WorldSense \citep{hong2026worldsense}. For RQ4, we evaluate perceptual capabilities on AVHBench \citep{sung2025avhbench}, MUSIC-AVQA v1/v2 \citep{li2022learning,liu2024tackling}, AVQA \citep{yang2022avqa}, and NExT-QA \citep{xiao2021next}. For RQ5, we jointly evaluate reasoning and perception using these benchmarks.

\paragraph{Baselines.}
We compare with the original Qwen2.5-Omni-7B \citep{Qwen2.5-Omni} and six representative omni-modal reasoning baselines: EchoInk-R1-7B \citep{xing2025echoink}, AVATAR-7B \citep{kulkarni2026avatar}, Omni-R1 \citep{zhong2026omni}, HumanSense \citep{qin2026humansense}, HumanOmniV2-7B \citep{yang2025humanomniv2}, and OmniVideo-7B \citep{cai2026omnivideo}.

\begin{table}[t]
\centering
\caption{Data types and sources for text-only and audio-visual training.}
\label{tab:training_data}
\fontsize{8}{9.5}\selectfont
\setlength{\tabcolsep}{2.5pt}
\renewcommand{\arraystretch}{1.04}
\setlength{\aboverulesep}{1.5pt}
\setlength{\belowrulesep}{1.5pt}

\begin{tabularx}{\linewidth}{@{}
    l
    l
    >{\raggedright\arraybackslash}X
    l
@{}}
\toprule
\textbf{Stage} &
\textbf{Data type} &
\textbf{Data sources} &
\textbf{RQ} \\
\midrule

\rowcolor{textgroup}[0pt][0pt]
\multicolumn{4}{@{}l@{}}{\textbf{Text-only training}} \\

SFT
& General reasoning
& SmallThoughts~\mbox{\citep{wu2025concise}};
Mixture-of-Thoughts~\mbox{\citep{openr1}};
Bespoke-Stratos-17k~\mbox{\citep{bespoke_stratos}};
OpenThoughts-114k~\mbox{\citep{guha2026openthoughts}};
OpenR1-Math-220k~\mbox{\citep{lozhkov2025openr1math220k}};
OpenMathReasoning~\mbox{\citep{moshkov2025aimo}}
& RQ2 \\
\addlinespace[1.5pt]

& Script-based CoT
& OmniVideo-100K~\mbox{\citep{cai2026omnivideo}}
& RQ1, 2, 4, 5 \\
\addlinespace[1.5pt]

& Contextual QA
& CoT-Drop~\mbox{\citep{liu2025instella}};
CoT-Collection~\mbox{\citep{kim2023cot}};
CoT-Scientific-RAG-Reasoning~\mbox{\citep{kumar2025cotscientificrag}}
& RQ2 \\

\midrule

GRPO
& Script-based MCQ
& OmniVideo-100K~\mbox{\citep{cai2026omnivideo}}
& RQ1, 2, 4, 5 \\
\addlinespace[1.5pt]

& Temporal QA
& Video-T3 QA~\mbox{\citep{li2025temporal}}
& RQ2 \\
\addlinespace[1.5pt]

& General \& math QA
& Nemotron-CrossThink~\mbox{\citep{akter2026nemotron}}
& RQ2 \\
\addlinespace[1.5pt]

& Contextual QA
& HotpotQA~\mbox{\citep{yang2018hotpotqa}};
MuSiQue~\mbox{\citep{trivedi2022musique}};
2WikiMultiHopQA~\mbox{\citep{ho2020constructing}}
& RQ2 \\

\midrule

SFT+GRPO
& Text-only LLM synthesis
& Qwen3.7-Max\footnotemark
& RQ3 \\

\midrule

\rowcolor{avgroup}[0pt][0pt]
\multicolumn{4}{@{}l@{}}{\textbf{Audio-visual training}} \\

SFT
& Audio-visual reasoning
& AVRT-20K~\mbox{\citep{araujo2026avrt}};
IntentTrain~\mbox{\citep{yang2025humanomniv2}}
& RQ1, 4, 5 \\

\midrule

GRPO
& Audio-visual MCQ
& OmniVideo-100K~\mbox{\citep{cai2026omnivideo}}
& RQ1, 4, 5 \\

\bottomrule
\end{tabularx}
\end{table}

\paragraph{Metrics.}
We report each benchmark's official metric on a 0--100 scale. \textbf{Macro Avg.} is the unweighted mean of scores, computed separately for reasoning and perception. \textbf{Geometric Mean Improvement} is the geometric mean of ratios to the corresponding base scores, minus one and expressed as a percentage. For RQ1 and RQ5, we report training input tokens and training compute in GPU-hours (number of GPUs $\times$ training time in hours) separately for each training stage.

\footnotetext{%
  \url{https://www.alibabacloud.com/help/en/model-studio/qwen3-7-max}%
}

\section{Findings}
\label{sec:findings}

We summarize the findings from our five RQs and their implications for Omni-LLM training:

\begin{itemize}
    \item \textbf{(RQ1)} Text-only reasoning training yields broad omni-modal reasoning gains with less training compute than the complete native audio-visual route.
    \item \textbf{(RQ2)} Script-based MCQs yield the strongest and broadest reasoning transfer.
    \item \textbf{(RQ3)} Training on data synthesized entirely by a text-only LLM also improves omni-modal reasoning, without native audio-visual data in construction or training.
    \item \textbf{(RQ4)} Text-only reasoning training incurs perceptual degradation.
    \item \textbf{(RQ5)} A text-centric training paradigm addresses the performance--compute trade-off by using text-only training for reasoning optimization and reduced-data native audio-visual refinement to maintain perception while retaining the main reasoning gains.
\end{itemize}

\subsection{RQ1: Reasoning Gains and Training Compute}
\label{sec:rq1}

\paragraph{Text-only reasoning training yields substantial and broad omni-modal reasoning gains.}
In Table~\ref{tab:rq1_main}, text-only SFT followed by GRPO raises Qwen2.5-Omni-7B's Macro Avg.\ from 39.80 to 49.39, a 9.59-point gain. Its Macro Avg.\ and geometric mean improvement of $+25.83\%$ both rank first. Trained only on text, it exceeds EchoInk-R1-7B and the complete native audio-visual route by 4.05 and 2.02 Macro Avg.\ points, respectively. It ranks first on all three SocialOmni subtasks, Video-MME-v2, and WorldSense, and second on Daily-Omni and OmniVideoBench. All nine scores improve over the base model, whereas every external baseline declines on at least one.


\begin{table}[t]
\centering
\caption{
Omni-modal reasoning performance and training compute.
Parenthesized values show absolute changes from the base model.
\textbf{Bold} and \underline{underlined} scores denote the best and second-best results, respectively.
}
\label{tab:rq1_main}
\begin{minipage}{\linewidth}
\centering
\setlength{\tabcolsep}{2.15pt}
\renewcommand{\arraystretch}{1.10}
\scriptsize

\begin{adjustbox}{max width=\linewidth}
\begin{tabular}{@{}l*{13}{c}@{}}
\toprule
\multirow{3}{*}{\textbf{Model}}
& \multicolumn{9}{c}{\textbf{Omni-modal Reasoning}}
& \multicolumn{2}{c}{\textbf{Aggregate}}
& \multicolumn{2}{c}{\textbf{Training Budget}} \\
\cmidrule(lr){2-10}\cmidrule(lr){11-12}\cmidrule(l){13-14}

& \multirow{2}{*}{\makecell{\textbf{Daily-}\\\textbf{Omni}}}
& \multirow{2}{*}{\makecell{\textbf{Omni-}\\\textbf{Cloze}}}
& \multirow{2}{*}{\makecell{\textbf{Omni}\\\textbf{Video}}}
& \multicolumn{3}{c}{\textbf{SocialOmni}}
& \multirow{2}{*}{\makecell{\textbf{Video-}\\\textbf{Holmes}}}
& \multirow{2}{*}{\makecell{\textbf{Video-}\\\textbf{MME-v2}}}
& \multirow{2}{*}{\makecell{\textbf{World}\\\textbf{Sense}}}
& \multirow{2}{*}{\makecell{\textbf{Macro}\\\textbf{Avg.}}}
& \multirow{2}{*}{\makecell{\textbf{Geo. Mean}\\\textbf{Improv.} (\%)}}
& \multirow{2}{*}{\makecell{\textbf{Input Tok.}\\\textbf{(B)} $\downarrow$}}
& \multirow{2}{*}{\makecell{\textbf{GPU-h}\\$\downarrow$}} \\
\cmidrule(lr){5-7}
& & & & {Who} & {When} & {How}
& & & & & & & \\
\midrule

Qwen2.5-Omni-7B
& 62.07 & 21.90 & 29.30
& 36.75 & 57.42 & 56.67
& 24.40 & 24.31 & 45.40
& 39.80 & 0.00
& \multicolumn{2}{c}{\textit{N/A}} \\

\addlinespace[1.4pt]
\rowcolor{groupbar}
\multicolumn{14}{@{}c@{}}{\textit{External Baselines}} \\

EchoInk-R1-7B~\citep{xing2025echoink}
& 61.07 & 27.30 & 33.50
& \underline{41.85} & 59.81 & 70.45
& 40.77 & 27.72 & 45.60
& 45.34 & +16.51
& 0.072 & 191.2 \\

AVATAR-7B~\citep{kulkarni2026avatar}
& 56.14 & 25.94 & 31.60
& 36.25 & 60.29 & 67.26
& \underline{45.10} & 20.47 & 46.00
& 43.23 & +9.37
& 0.419 & 385.9 \\

Omni-R1~\citep{zhong2026omni}
& 55.56 & 26.28 & 32.10
& 33.30 & 62.68 & 72.55
& 32.12 & 25.04 & 41.40
& 42.34 & +7.03
& 0.055 & 576.0 \\

HumanSense~\citep{qin2026humansense}
& 56.31 & 26.98 & 31.70
& 38.70 & \underline{66.51} & \underline{75.86}
& 33.15 & 24.33 & 42.20
& 43.97 & +10.69
& 0.208 & 190.3 \\

HumanOmniV2-7B~\citep{yang2025humanomniv2}
& 58.47 & \textbf{29.38} & 30.50
& 40.20 & 57.89 & 66.03
& 42.90 & 24.93 & 47.10
& 44.16 & +13.65
& 4.506 & 2164.1 \\

OmniVideo-7B~\citep{cai2026omnivideo}
& \textbf{68.34} & 25.40 & \textbf{38.30}
& 35.45 & 41.15 & 50.40
& 38.00 & 24.54 & 45.10
& 40.74 & +5.38
& 1.511 & 201.0 \\

\addlinespace[1.4pt]
\rowcolor{groupbar}
\multicolumn{14}{@{}c@{}}{\textit{Training Variants}} \\

\rowcolor{avgroup}
Audio-visual SFT
& 54.64 & \underline{28.32} & 32.90
& 37.25 & 63.64 & 56.76
& 36.74 & 26.96 & 43.40
& 42.29 & +9.67
& 0.092 & 12.2 \\

\rowcolor{avgroup}
\hspace{1.45em}$+$ GRPO
& 62.99 & 26.48 & 36.70
& 41.40 & 64.59 & 69.76
& \textbf{46.65} & \underline{30.20} & \underline{47.60}
& \underline{47.37} & \underline{+21.96}
& 1.214 & 642.3 \\

\addlinespace[1.0pt]
\rowcolor{textgroup}
\textbf{Text-only SFT}
& 52.72 & 25.25 & 31.40
& 39.30 & 53.59 & 66.07
& 27.71 & 22.34 & 40.90
& 39.92 & +1.55
& 0.043 & 10.2 \\

\rowcolor{textgroup}
\hspace{1.45em}$+$ \textbf{GRPO (Ours)}
& \underline{65.75} & 27.17 & \underline{37.00}
& \textbf{41.95} & \textbf{68.90} & \textbf{79.93}
& 44.58 & \textbf{30.66} & \textbf{48.60}
& \textbf{49.39} & \textbf{+25.83}
& 0.463 & 274.0 \\

& {\tiny\bfseries\boldmath\color{deltapos} $(+3.68)$}
& {\tiny\bfseries\boldmath\color{deltapos} $(+5.27)$}
& {\tiny\bfseries\boldmath\color{deltapos} $(+7.70)$}
& {\tiny\bfseries\boldmath\color{deltapos} $(+5.20)$}
& {\tiny\bfseries\boldmath\color{deltapos} $(+11.48)$}
& {\tiny\bfseries\boldmath\color{deltapos} $(+23.26)$}
& {\tiny\bfseries\boldmath\color{deltapos} $(+20.18)$}
& {\tiny\bfseries\boldmath\color{deltapos} $(+6.35)$}
& {\tiny\bfseries\boldmath\color{deltapos} $(+3.20)$}
& {\tiny\bfseries\boldmath\color{deltapos} $(+9.59)$}
& & & \\

\bottomrule
\end{tabular}
\end{adjustbox}
\end{minipage}
\end{table}


\begin{table}[t]
\centering
\caption{
Generalization of text-only reasoning training across model scales and families.
Parenthesized values show absolute changes from the corresponding base model.
}
\label{tab:rq1_generalization}
\setlength{\tabcolsep}{2.15pt}
\renewcommand{\arraystretch}{1.10}
\scriptsize

\begin{adjustbox}{max width=\linewidth}
\begin{tabular}{@{}l*{13}{c}@{}}
\toprule
\multirow{3}{*}{\textbf{Model}}
& \multicolumn{9}{c}{\textbf{Omni-modal Reasoning}}
& \multicolumn{2}{c}{\textbf{Aggregate}}
& \multicolumn{2}{c}{\textbf{Training Budget}} \\
\cmidrule(lr){2-10}\cmidrule(lr){11-12}\cmidrule(l){13-14}
& \multirow{2}{*}{\makecell{\textbf{Daily-}\\\textbf{Omni}}}
& \multirow{2}{*}{\makecell{\textbf{Omni-}\\\textbf{Cloze}}}
& \multirow{2}{*}{\makecell{\textbf{Omni}\\\textbf{Video}}}
& \multicolumn{3}{c}{\textbf{SocialOmni}}
& \multirow{2}{*}{\makecell{\textbf{Video-}\\\textbf{Holmes}}}
& \multirow{2}{*}{\makecell{\textbf{Video-}\\\textbf{MME-v2}}}
& \multirow{2}{*}{\makecell{\textbf{World}\\\textbf{Sense}}}
& \multirow{2}{*}{\makecell{\textbf{Macro}\\\textbf{Avg.}}}
& \multirow{2}{*}{\makecell{\textbf{Geo. Mean}\\\textbf{Improv.} (\%)}}
& \multirow{2}{*}{\makecell{\textbf{Input Tok.}\\\textbf{(B)} $\downarrow$}}
& \multirow{2}{*}{\makecell{\textbf{GPU-h}\\$\downarrow$}} \\
\cmidrule(lr){5-7}
& & & & {Who} & {When} & {How} & & & & & & & \\
\midrule

Qwen2.5-Omni-3B
& 56.22 & 23.51 & 27.30
& 36.50 & 50.24 & 51.23
& 25.26 & 21.35 & 43.50
& 37.23 & 0.00
& \multicolumn{2}{c}{\textit{N/A}} \\

\rowcolor{textgroup}
\textbf{Text-only SFT}
& 51.04 & 23.52 & 32.20
& 32.55 & 55.50 & 58.93
& 32.01 & 20.46 & 40.00
& 38.47 & +3.44
& 0.128 & 10.1 \\

\rowcolor{textgroup}
\hspace{1.15em}$+$ \textbf{GRPO}
& 64.16 & 24.40 & 36.00
& 37.10 & 57.42 & 71.88
& 42.95 & 25.17 & 45.30
& 44.93 & +20.39
& 0.463 & 259.3 \\

& {\tiny\bfseries\boldmath\color{deltapos} $(+7.94)$}
& {\tiny\bfseries\boldmath\color{deltapos} $(+0.89)$}
& {\tiny\bfseries\boldmath\color{deltapos} $(+8.70)$}
& {\tiny\bfseries\boldmath\color{deltapos} $(+0.60)$}
& {\tiny\bfseries\boldmath\color{deltapos} $(+7.18)$}
& {\tiny\bfseries\boldmath\color{deltapos} $(+20.65)$}
& {\tiny\bfseries\boldmath\color{deltapos} $(+17.69)$}
& {\tiny\bfseries\boldmath\color{deltapos} $(+3.82)$}
& {\tiny\bfseries\boldmath\color{deltapos} $(+1.80)$}
& {\tiny\bfseries\boldmath\color{deltapos} $(+7.70)$}
& & & \\

\midrule
Gemma-4-E4B-IT
& 50.54 & 25.16 & 28.10
& 35.30 & 52.63 & 69.17
& 32.55 & 21.81 & 35.80
& 39.01 & 0.00
& \multicolumn{2}{c}{\textit{N/A}} \\

\rowcolor{textgroup}
\textbf{Text-only SFT}
& 48.37 & 23.86 & 30.60
& 33.05 & 59.33 & 76.47
& 31.95 & 20.37 & 35.20
& 39.91 & +0.43
& 0.042 & 12.9 \\

\rowcolor{textgroup}
\hspace{1.15em}$+$ \textbf{GRPO}
& 57.31 & 25.77 & 33.10
& 36.40 & 61.24 & 79.73
& 42.84 & 21.30 & 39.20
& 44.10 & +11.50
& 0.572 & 289.0 \\

& {\tiny\bfseries\boldmath\color{deltapos} $(+6.77)$}
& {\tiny\bfseries\boldmath\color{deltapos} $(+0.61)$}
& {\tiny\bfseries\boldmath\color{deltapos} $(+5.00)$}
& {\tiny\bfseries\boldmath\color{deltapos} $(+1.10)$}
& {\tiny\bfseries\boldmath\color{deltapos} $(+8.61)$}
& {\tiny\bfseries\boldmath\color{deltapos} $(+10.56)$}
& {\tiny\bfseries\boldmath\color{deltapos} $(+10.29)$}
& {\tiny\bfseries\boldmath\color{deltaneg} $(-0.51)$}
& {\tiny\bfseries\boldmath\color{deltapos} $(+3.40)$}
& {\tiny\bfseries\boldmath\color{deltapos} $(+5.09)$}
& & & \\

\midrule

Gemma-4-E2B-IT
& 48.20 & 26.45 & 28.00
& 34.35 & 51.67 & 66.18
& 30.00 & 19.56 & 33.50
& 37.55 & 0.00
& \multicolumn{2}{c}{\textit{N/A}} \\

\rowcolor{textgroup}
\textbf{Text-only SFT}
& 47.45 & 22.76 & 29.00
& 33.05 & 57.89 & 72.75
& 28.14 & 18.69 & 35.80
& 38.39 & -0.05
& 0.042 & 9.8 \\

\rowcolor{textgroup}
\hspace{1.15em}$+$ \textbf{GRPO}
& 51.96 & 24.03 & 30.50
& 34.15 & 62.20 & 73.28
& 33.70 & 20.09 & 36.90
& 40.76 & +6.73
& 0.565 & 215.2 \\

& \posd{3.76}
& \negd{2.42}
& \posd{2.50}
& \negd{0.20}
& \posd{10.53}
& \posd{7.10}
& \posd{3.70}
& \posd{0.53}
& \posd{3.40}
& \posd{3.21}
& & & \\

\midrule

Qwen3-Omni-30B
& 71.85 & 33.50 & 38.40
& 69.25 & 63.64 & 46.35
& 50.40 & 27.00 & 54.00
& 50.49 & 0.00
& \multicolumn{2}{c}{\textit{N/A}} \\

\rowcolor{textgroup}
\textbf{Text-only SFT}
& 64.66 & 36.08 & 41.30
& 66.15 & 59.81 & 73.28
& 54.00 & 29.62 & 52.74
& 53.07 & +6.06
& 0.135 & 31.1 \\

& {\tiny\bfseries\boldmath\color{deltaneg} $(-7.19)$}
& {\tiny\bfseries\boldmath\color{deltapos} $(+2.58)$}
& {\tiny\bfseries\boldmath\color{deltapos} $(+2.90)$}
& {\tiny\bfseries\boldmath\color{deltaneg} $(-3.10)$}
& {\tiny\bfseries\boldmath\color{deltaneg} $(-3.83)$}
& {\tiny\bfseries\boldmath\color{deltapos} $(+26.93)$}
& {\tiny\bfseries\boldmath\color{deltapos} $(+3.60)$}
& {\tiny\bfseries\boldmath\color{deltapos} $(+2.62)$}
& {\tiny\bfseries\boldmath\color{deltaneg} $(-1.26)$}
& {\tiny\bfseries\boldmath\color{deltapos} $(+2.58)$}
& & & \\

\bottomrule
\end{tabular}
\end{adjustbox}
\end{table}

\paragraph{The reasoning gains extend across model scales and families.} In Table~\ref{tab:rq1_generalization}, text-only SFT followed by GRPO raises Qwen2.5-Omni-3B's Macro Avg.\ from 37.23 to 44.93, with a geometric mean improvement of $+20.39\%$. It improves all nine scores over the base model, including gains of 20.65 and 17.69 points on SocialOmni How and Video-Holmes, respectively. On Gemma-4-E4B-IT and Gemma-4-E2B-IT, Macro Avg.\ rises by 5.09 and 3.21 points, and eight and seven scores improve, respectively. For the latter, GRPO improves all nine scores over SFT despite uneven SFT gains. Even on the stronger Qwen3-Omni-30B-A3B-Instruct, text-only SFT alone raises the Macro Avg.\ from 50.49 to 53.07. These results support text-only training across scales and families.

\paragraph{Text-only reasoning training achieves a superior performance--compute trade-off.}
The complete text-only route in Table~\ref{tab:rq1_main} improves Macro Avg.\ (49.39 versus 47.37) with fewer GPU-hours (284.17 versus 654.52) than the native audio-visual route. Across SFT and GRPO, it processes 0.505B versus 1.306B training input tokens. This reduces input tokens by 61.3\% and training compute by 56.6\%. It achieves the highest Macro Avg.\ among compared methods with fewer GPU-hours than measured for AVATAR-7B, Omni-R1, and HumanOmniV2-7B. Thus, text-only training outperforms the complete native audio-visual route with less training compute.

\subsection{RQ2: Choice of Training Data}
\label{sec:rq2}

\begin{table}[t]
\centering
\caption{
Effect of training data choice on omni-modal reasoning.
\textbf{Bold} and \underline{underlined} scores denote the best and second-best results, respectively.
}
\label{tab:rq2_data}
\setlength{\tabcolsep}{7.20pt}
\renewcommand{\arraystretch}{1.10}
\scriptsize

\begin{adjustbox}{max width=\linewidth}
\begin{tabular}{@{}l*{11}{c}@{}}
\toprule
\multirow{3}{*}{\textbf{Training Configuration}}
& \multicolumn{9}{c}{\textbf{Omni-modal Reasoning}}
& \multicolumn{2}{c}{\textbf{Aggregate}} \\
\cmidrule(lr){2-10}
\cmidrule(l){11-12}

& \multirow{2}{*}{\makecell{\textbf{Daily-}\\\textbf{Omni}}}
& \multirow{2}{*}{\makecell{\textbf{Omni-}\\\textbf{Cloze}}}
& \multirow{2}{*}{\makecell{\textbf{Omni}\\\textbf{Video}}}
& \multicolumn{3}{c}{\textbf{SocialOmni}}
& \multirow{2}{*}{\makecell{\textbf{Video-}\\\textbf{Holmes}}}
& \multirow{2}{*}{\makecell{\textbf{Video-}\\\textbf{MME-v2}}}
& \multirow{2}{*}{\makecell{\textbf{World}\\\textbf{Sense}}}
& \multirow{2}{*}{\makecell{\textbf{Macro}\\\textbf{Avg.}}}
& \multirow{2}{*}{\makecell{\textbf{Geo. Mean}\\\textbf{Improv.} (\%)}} \\
\cmidrule(lr){5-7}
& & & & {Who} & {When} & {How}
& & & & & \\
\midrule

Qwen2.5-Omni-7B
& 62.07 & 21.90 & 29.30
& 36.75 & 57.42 & 56.67
& 24.40 & 24.31 & 45.40
& 39.80 & 0.00 \\

\midrule

\textbf{SFT:} General reasoning
& 55.72 & 26.58 & 31.40
& 37.35 & 56.94 & 69.02
& 37.51 & 25.49 & 45.90
& 42.88 & +9.88 \\

\hspace{1.15em}$+$ \textbf{GRPO:} Script-based MCQ
& \textbf{66.58}
& \textbf{28.56}
& \underline{36.90}
& \underline{41.45}
& \underline{63.64}
& 75.42
& \textbf{47.25}
& \underline{30.41}
& \textbf{50.00}
& \underline{48.91}
& \underline{+25.67} \\

\hspace{1.15em}$+$ \textbf{GRPO:} Temporal QA
& 55.22 & 27.01 & 34.00
& 38.60 & 57.89 & \underline{76.09}
& 38.54 & 25.75 & 44.90
& 44.22 & +12.96 \\

\hspace{1.15em}$+$ \textbf{GRPO:} General \& math QA
& 55.56 & 26.88 & 33.70
& 38.65 & \underline{63.64} & 74.59
& 38.16 & 25.92 & 44.70
& 44.64 & +13.73 \\

\midrule

\rowcolor{textgroup}
\textbf{SFT:} Script-based CoT (MCQ)
& 52.72 & 25.25 & 31.40
& 39.30 & 53.59 & 66.07
& 27.71 & 22.34 & 40.90
& 39.92 & +1.55 \\

\rowcolor{textgroup}
\hspace{1.15em}$+$ \textbf{GRPO:} Script-based MCQ
& \underline{65.75}
& \underline{27.17}
& \textbf{37.00}
& \textbf{41.95}
& \textbf{68.90}
& \textbf{79.93}
& \underline{44.58}
& \textbf{30.66}
& \underline{48.60}
& \textbf{49.39}
& \textbf{+25.83} \\

\hspace{1.15em}$+$ \textbf{GRPO:} Temporal QA
& 57.06 & 25.88 & 33.57
& 37.90 & 57.42 & 71.08
& 36.25 & 22.81 & 42.09
& 42.67 & +8.48 \\

\hspace{1.15em}$+$ \textbf{GRPO:} General \& math QA
& 55.81 & 25.97 & 31.56
& 37.55 & 55.50 & 67.93
& 35.49 & 25.22 & 42.72
& 41.97 & +7.58 \\

\midrule

\textbf{SFT:} Contextual QA
& 52.88 & 26.46 & 32.80
& 38.25 & 54.07 & 70.92
& 27.82 & 21.44 & 39.90
& 40.50 & +2.50 \\

\hspace{1.15em}$+$ \textbf{GRPO:} Contextual QA
& 55.14 & 26.47 & 31.20
& 37.15 & 59.81 & 73.98
& 35.44 & 25.24 & 42.30
& 42.97 & +9.20 \\

\bottomrule
\end{tabular}
\end{adjustbox}
\end{table}

\paragraph{Script-based MCQ yields stronger omni-modal reasoning transfer.}
In Table~\ref{tab:rq2_data}, GRPO on Script-based MCQs achieves the highest Macro Avg.\ after SFT on either General reasoning data or Script-based MCQs with CoT supervision, reaching 48.91 and 49.39, respectively. Contextual QA in both phases yields 42.97. Both MCQ configurations improve all nine scores over the base model, whereas other complete configurations improve the Macro Avg.\ but not all scores. Tracking entities and event states and integrating evidence across segments may explain these broader gains in omni-modal reasoning. Further results on data mixtures appear in Appendix~\ref{sec:appendix_training_data}.

\begin{figure}[ht]
    \centering
    \includegraphics[width=0.89\linewidth]{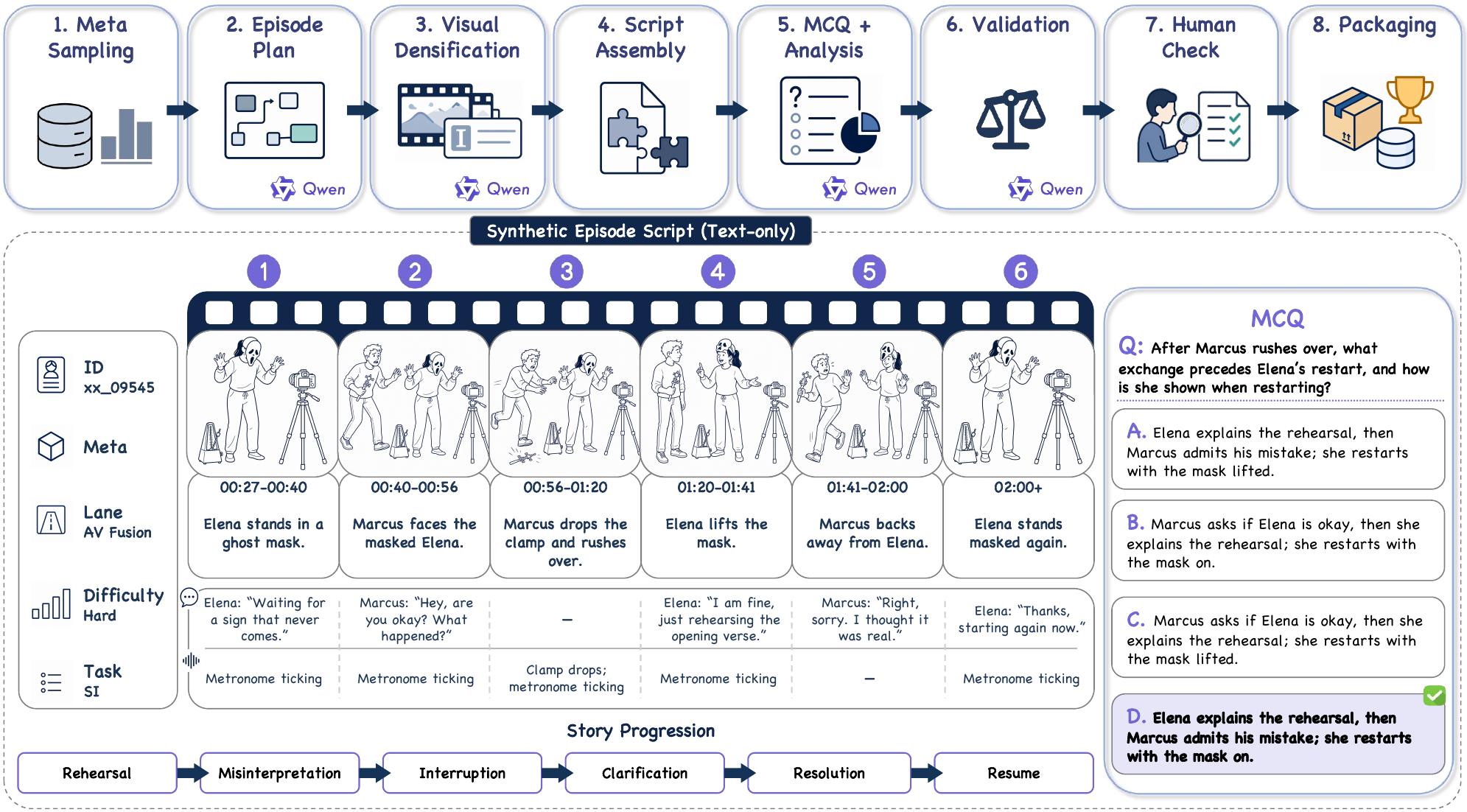}
    \caption{
    Text-only LLM synthesis of scripts, MCQs, and reasoning explanations, without audio-visual data in construction or training.
    The illustration shows a fictional script, not generated video.
    }
    \label{fig:rq3_synthesis_pipeline}
\end{figure}

\subsection{RQ3: Synthetic Training Data}
\label{sec:rq3}

\begin{figure}[t]
    \centering
    \includegraphics[width=0.89\linewidth]{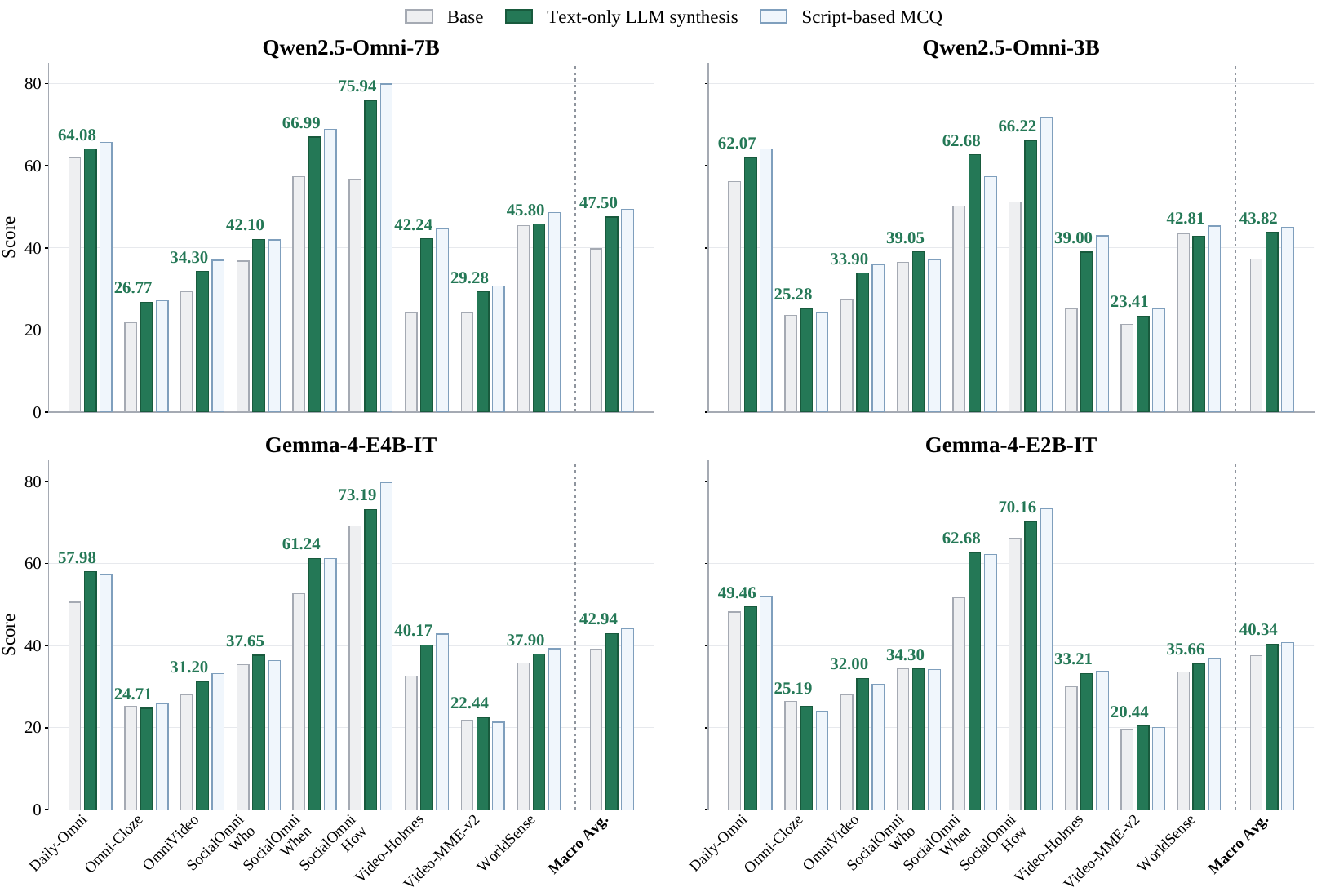}
    \caption{
    Omni-modal reasoning with training data synthesized entirely by a text-only LLM
    for both SFT and GRPO.
    Results cover four models across different scales and families.
    }
    \label{fig:rq3_synthetic_results}
\end{figure}

\paragraph{Text-only LLM synthesis extends text-only training to data construction.}
Building on RQ1, we test text-only reasoning training on SFT and GRPO data synthesized entirely by a text-only LLM. Neither data construction nor training uses any native audio-visual data. Figure~\ref{fig:rq3_synthesis_pipeline} shows generation from specified scenarios, task types, and modality requirements. The LLM creates fictional episode scripts with visual descriptions, speech, and non-speech sounds, maintaining entity and temporal consistency across segments. MCQs require entity tracking, event-state updates, and cross-segment evidence integration; joint audio-visual questions require complementary audio and visual evidence. Scripts, MCQs, and reasoning explanations undergo validation and human review.

\paragraph{Text-only LLM synthesis yields omni-modal reasoning gains.}
In Figure~\ref{fig:rq3_synthetic_results}, training on synthesized text yields a 7.70-point Macro Avg.\ gain for Qwen2.5-Omni-7B, from 39.80 to 47.50. Gains on Qwen2.5-Omni-3B, Gemma-4-E4B-IT, and Gemma-4-E2B-IT are 6.59, 3.93, and 2.79 points, respectively. The Qwen models improve nine and eight scores over their base models and remain within 1.89 and 1.11 Macro Avg.\ points of Script-based MCQ. Crucially, these gains require no native audio-visual data in construction or training. This opens a path to scaling up reasoning training data without collecting or annotating corresponding audio-visual samples.

\begin{table}[t]
\centering
\caption{
Perceptual effects of text-only and audio-visual training.
Parenthesized values show absolute changes from the base model.
}
\label{tab:rq4_perception}
\begin{minipage}{\linewidth}
\centering
\setlength{\tabcolsep}{2.55pt}
\renewcommand{\arraystretch}{1.06}
\scriptsize

\begin{adjustbox}{max width=\linewidth}
\begin{tabular}{@{}l*{9}{c}@{}}
\toprule
\multirow{3}{*}{\textbf{Model}}
& \multicolumn{7}{c}{\textbf{Perceptual Capabilities}}
& \multicolumn{2}{c}{\textbf{Aggregate}} \\
\cmidrule(lr){2-8}
\cmidrule(l){9-10}

& \multicolumn{3}{c}{\textbf{AVHBench}}
& \multicolumn{2}{c}{\textbf{MUSIC-AVQA}}
& \textbf{AVQA}
& \textbf{NExT-QA}
& \multirow{2}{*}{\makecell{\textbf{Macro}\\\textbf{Avg.}}}
& \multirow{2}{*}{\makecell{\textbf{Geo. Mean}\\\textbf{Improv.} (\%)}} \\
\cmidrule(lr){2-4}
\cmidrule(lr){5-6}
\cmidrule(lr){7-7}
\cmidrule(lr){8-8}

& {Overall}
& {\makecell{Audio-driven\\Video Hallucination}}
& {\makecell{Video-driven\\Audio Hallucination}}
& {v1}
& {v2}
& {\makecell{Which / Come From /\\
           Happening / Where}}
& {Descriptive}
& & \\
\midrule

Qwen2.5-Omni-7B
& 75.55
& 80.70
& 73.00
& 71.22
& 69.10
& 88.79
& 82.21
& 77.37
& 0.00 \\

\addlinespace[1.0pt]
\rowcolor{avgroup}
Audio-visual SFT
& 78.12 & 82.60 & 75.91 & 75.91 & 74.62 & 90.70 & 83.13 & 80.50 & +4.22 \\

\rowcolor{avgroup}
\hspace{1.15em}$+$ GRPO
& 78.31
& 81.87
& 76.55
& 73.15
& 70.32
& 90.58
& 82.99
& 79.07
& +2.21 \\

&
{\tiny\bfseries\boldmath\color{deltapos} $(+2.76)$}
& {\tiny\bfseries\boldmath\color{deltapos} $(+1.17)$}
& {\tiny\bfseries\boldmath\color{deltapos} $(+3.55)$}
& {\tiny\bfseries\boldmath\color{deltapos} $(+1.93)$}
& {\tiny\bfseries\boldmath\color{deltapos} $(+1.22)$}
& {\tiny\bfseries\boldmath\color{deltapos} $(+1.79)$}
& {\tiny\bfseries\boldmath\color{deltapos} $(+0.78)$}
& {\tiny\bfseries\boldmath\color{deltapos} $(+1.70)$}
& \\

\addlinespace[0.8pt]

\rowcolor{textgroup}
\textbf{Text-only SFT}
& 75.78 & 80.30 & 73.55 & 65.09 & 62.93 & 83.75 & 79.22 & 73.35 & -5.37 \\

\rowcolor{textgroup}
\hspace{1.15em}$+$ \textbf{GRPO (Ours)}
& 73.94
& 79.42
& 71.23
& 69.60
& 68.99
& 87.68
& 81.85
& 76.41
& -1.25 \\

&
{\tiny\bfseries\boldmath\color{deltaneg} $(-1.61)$}
& {\tiny\bfseries\boldmath\color{deltaneg} $(-1.28)$}
& {\tiny\bfseries\boldmath\color{deltaneg} $(-1.77)$}
& {\tiny\bfseries\boldmath\color{deltaneg} $(-1.62)$}
& {\tiny\bfseries\boldmath\color{deltaneg} $(-0.11)$}
& {\tiny\bfseries\boldmath\color{deltaneg} $(-1.11)$}
& {\tiny\bfseries\boldmath\color{deltaneg} $(-0.36)$}
& {\tiny\bfseries\boldmath\color{deltaneg} $(-0.96)$}
& \\

\bottomrule
\end{tabular}
\end{adjustbox}
\end{minipage}
\end{table}

\subsection{RQ4: Perceptual Effects}
\label{sec:rq4}

\paragraph{Text-only reasoning training incurs perceptual degradation.}
In Table~\ref{tab:rq4_perception}, text-only SFT followed by GRPO reduces Qwen2.5-Omni-7B's perceptual Macro Avg.\ from 77.37 to 76.41, with a 1.25\% decrease in the geometric mean of scores. All seven perceptual scores decline by 0.11--1.77 points. In contrast, native audio-visual SFT followed by GRPO improves all seven scores and raises the Macro Avg.\ by 1.70 points to 79.07. The text-only route's 9.59-point reasoning gain in RQ1 accompanies a 0.96-point perceptual Macro Avg.\ decline. This pattern is consistent with the partial decoupling of perception and reasoning suggested by our preliminary study. Nevertheless, text-only reasoning training does not fully replace native audio-visual training for preserving perception.

\subsection{RQ5: Training Paradigm}
\label{sec:rq5}

\begin{table}[t]
\centering
\begin{minipage}{\linewidth}
\centering
\captionsetup{font=small,justification=justified,singlelinecheck=false,skip=5pt}
\caption{
Reasoning, perception, and training budgets under staged training.
A$\to$V and V$\to$A denote audio-driven video and video-driven audio hallucination.
Parentheses show absolute changes from the base model. Budgets are per audio-visual stage; reduced-data budgets are shared across initializations.
}
\label{tab:rq5_staged}
\setlength{\tabcolsep}{2.05pt}
\renewcommand{\arraystretch}{1.08}
\scriptsize

\begin{adjustbox}{max width=\linewidth}
\begin{tabular}{@{}ll*{11}{c}@{}}
\toprule
\multicolumn{13}{c}{\textbf{(a) Omni-modal Reasoning}} \\
\midrule
\multicolumn{2}{@{}l}{\textbf{Training Configuration}}
& \multirow{2}{*}{\makecell{\textbf{Daily-}\\\textbf{Omni}}}
& \multirow{2}{*}{\makecell{\textbf{Omni-}\\\textbf{Cloze}}}
& \multirow{2}{*}{\makecell{\textbf{Omni}\\\textbf{Video}}}
& \multicolumn{3}{c}{\textbf{SocialOmni}}
& \multirow{2}{*}{\makecell{\textbf{Video-}\\\textbf{Holmes}}}
& \multirow{2}{*}{\makecell{\textbf{Video-}\\\textbf{MME-v2}}}
& \multirow{2}{*}{\makecell{\textbf{World}\\\textbf{Sense}}}
& \multirow{2}{*}{\makecell{\textbf{Macro}\\\textbf{Avg.}}}
& \multirow{2}{*}{\makecell{\textbf{Geo. Mean}\\\textbf{Improv.} (\%)}} \\
\cmidrule(lr){1-2}\cmidrule(lr){6-8}
\textbf{Text-only} & \textbf{Audio-visual} & & & & {Who} & {When} & {How} & & & & & \\
\midrule

\multicolumn{2}{@{}l}{Qwen2.5-Omni-7B}
& 62.07 & 21.90 & 29.30 & 36.75 & 57.42 & 56.67 & 24.40 & 24.31 & 45.40 & 39.80 & 0.00 \\

\addlinespace[1.0pt]
\rowcolor{groupbar}
\multicolumn{13}{@{}c@{}}{\textit{Reduced-data Audio-visual SFT}} \\
-- & \cellcolor{avgroup}SFT
& 51.38 & 25.86 & 31.90 & 37.90 & 56.46 & 68.52 & 34.08 & 20.69 & 40.10 & 40.77 & +3.56 \\
\cellcolor{textgroup}SFT & \cellcolor{avgroup}SFT
& 55.14 & 25.74 & 37.40 & 37.30 & 59.81 & 71.59 & 35.49 & 23.38 & 41.60 & 43.05 & +9.60 \\
\cellcolor{textgroup}SFT + GRPO & \cellcolor{avgroup}SFT
& 65.33 & 27.06 & 37.40 & 42.10 & 63.64 & 70.56 & 45.18 & 28.09 & 47.40 & 47.42 & +21.71 \\

\addlinespace[1.0pt]
\rowcolor{groupbar}
\multicolumn{13}{@{}c@{}}{\textit{Reduced-data Audio-visual GRPO}} \\
\cellcolor{textgroup}SFT & \cellcolor{avgroup}GRPO
& 62.74 & 26.75 & 37.47 & 40.80 & 66.51 & 72.35 & 43.22 & 27.98 & 47.10 & 47.21 & +20.81 \\
\cellcolor{textgroup}SFT + GRPO & \cellcolor{avgroup}GRPO
& 64.75 & 27.08 & 37.17 & 41.15 & 66.51 & 75.00 & 45.45 & 30.16 & 49.02 & 48.48 & +24.15 \\

&
& {\tiny\bfseries\boldmath\color{deltapos} $(+2.68)$}
& {\tiny\bfseries\boldmath\color{deltapos} $(+5.18)$}
& {\tiny\bfseries\boldmath\color{deltapos} $(+7.87)$}
& {\tiny\bfseries\boldmath\color{deltapos} $(+4.40)$}
& {\tiny\bfseries\boldmath\color{deltapos} $(+9.09)$}
& {\tiny\bfseries\boldmath\color{deltapos} $(+18.33)$}
& {\tiny\bfseries\boldmath\color{deltapos} $(+21.05)$}
& {\tiny\bfseries\boldmath\color{deltapos} $(+5.85)$}
& {\tiny\bfseries\boldmath\color{deltapos} $(+3.62)$}
& {\tiny\bfseries\boldmath\color{deltapos} $(+8.67)$}
& \\

\midrule
\multicolumn{13}{c}{\textbf{(b) Perceptual Capabilities}} \\
\midrule
\multicolumn{2}{@{}l}{\textbf{Training Configuration}}
& \multicolumn{5}{c}{\textbf{AVHBench}}
& \multicolumn{2}{c}{\textbf{MUSIC-AVQA}}
& \textbf{AVQA}
& \textbf{NExT-QA}
& \multirow{2}{*}{\makecell{\textbf{Macro}\\\textbf{Avg.}}}
& \multirow{2}{*}{\makecell{\textbf{Geo. Mean}\\\textbf{Improv.} (\%)}} \\[-0.8pt]
\cmidrule(lr){1-2}\cmidrule(lr){3-7}\cmidrule(lr){8-9}\cmidrule(lr){10-10}\cmidrule(lr){11-11}
\noalign{\vskip -0.6pt}
\textbf{Text-only} & \textbf{Audio-visual}
& {Overall}
& \multicolumn{2}{c}{{A$\to$V}}
& \multicolumn{2}{c}{{V$\to$A}}
& {v1}
& {v2}
& {\makecell{Which / Come From /\\[-1pt]Happening / Where}}
& {Descriptive}
& & \\[-0.6pt]
\midrule

\multicolumn{2}{@{}l}{Qwen2.5-Omni-7B}
& 75.55 & \multicolumn{2}{c}{80.70} & \multicolumn{2}{c}{73.00}
& 71.22 & 69.10 & 88.79 & 82.21 & 77.37 & 0.00 \\

\addlinespace[1.0pt]
\rowcolor{groupbar}
\multicolumn{13}{@{}c@{}}{\textit{Reduced-data Audio-visual SFT}} \\
-- & \cellcolor{avgroup}SFT
& 76.25 & \multicolumn{2}{c}{80.83} & \multicolumn{2}{c}{73.98}
& 74.50 & 71.64 & 90.67 & 83.35 & 79.28 & +2.53 \\
\cellcolor{textgroup}SFT & \cellcolor{avgroup}SFT
& 76.31 & \multicolumn{2}{c}{81.10} & \multicolumn{2}{c}{73.94}
& 74.41 & 71.67 & 90.64 & 83.20 & 79.25 & +2.49 \\
\cellcolor{textgroup}SFT + GRPO & \cellcolor{avgroup}SFT
& 74.87 & \multicolumn{2}{c}{80.02} & \multicolumn{2}{c}{72.32}
& 73.14 & 70.42 & 90.39 & 83.84 & 78.53 & +1.49 \\

\addlinespace[1.0pt]
\rowcolor{groupbar}
\multicolumn{13}{@{}c@{}}{\textit{Reduced-data Audio-visual GRPO}} \\
\cellcolor{textgroup}SFT & \cellcolor{avgroup}GRPO
& 75.84 & \multicolumn{2}{c}{80.65} & \multicolumn{2}{c}{73.46}
& 74.19 & 71.30 & 89.14 & 82.70 & 78.63 & +1.73 \\
\cellcolor{textgroup}SFT + GRPO & \cellcolor{avgroup}GRPO
& 76.06 & \multicolumn{2}{c}{81.60} & \multicolumn{2}{c}{73.32}
& 73.22 & 70.49 & 89.69 & 83.06 & 78.50 & +1.51 \\

&
& {\tiny\bfseries\boldmath\color{deltapos} $(+0.51)$}
& \multicolumn{2}{c}{\tiny\bfseries\boldmath\color{deltapos} $(+0.90)$}
& \multicolumn{2}{c}{\tiny\bfseries\boldmath\color{deltapos} $(+0.32)$}
& {\tiny\bfseries\boldmath\color{deltapos} $(+2.00)$}
& {\tiny\bfseries\boldmath\color{deltapos} $(+1.39)$}
& {\tiny\bfseries\boldmath\color{deltapos} $(+0.90)$}
& {\tiny\bfseries\boldmath\color{deltapos} $(+0.85)$}
& {\tiny\bfseries\boldmath\color{deltapos} $(+1.13)$}
& \\
\midrule
\multicolumn{13}{c}{\textbf{(c) Training Budget}} \\
\midrule
\multicolumn{2}{@{}l}{\multirow{2}{*}{\textbf{Metric}}}
& \multicolumn{6}{c}{\textbf{Full-data Audio-visual}}
& \multicolumn{5}{c}{\textbf{Reduced-data Audio-visual}} \\
\cmidrule(lr){3-8}\cmidrule(l){9-13}
\multicolumn{2}{@{}l}{}
& \multicolumn{3}{c}{\textbf{SFT}}
& \multicolumn{3}{c}{\textbf{GRPO}}
& \multicolumn{2}{c}{\textbf{SFT}}
& \multicolumn{3}{c}{\textbf{GRPO}} \\
\midrule
\multicolumn{2}{@{}l}{Input Tok. (B) $\downarrow$}
& \multicolumn{3}{c}{0.092}
& \multicolumn{3}{c}{1.214}
& \multicolumn{2}{c}{0.002}
& \multicolumn{3}{c}{0.122} \\
\multicolumn{2}{@{}l}{GPU-h $\downarrow$}
& \multicolumn{3}{c}{12.2}
& \multicolumn{3}{c}{642.3}
& \multicolumn{2}{c}{0.4}
& \multicolumn{3}{c}{82.7} \\
\bottomrule
\end{tabular}
\end{adjustbox}
\end{minipage}
\end{table}

\paragraph{RQ1--RQ4 motivate a text-centric training paradigm.}
Building on the reasoning gains in RQ1--RQ3 and perceptual degradation in RQ4, we investigate text-centric post-training. Once reasoning has been strengthened through text-only training, we test whether reduced-data native audio-visual training is sufficient to preserve perception. Specifically, we apply audio-visual SFT or GRPO after either text-only SFT or the complete text-only SFT--GRPO pipeline, with reduced-data audio-visual SFT alone as a control. Table~\ref{tab:rq5_staged} reports reasoning, perception, and training budgets.

\paragraph{Staged training improves reasoning while maintaining perception.}
Table~\ref{tab:rq5_staged} shows higher reasoning Macro Avg.\ scores from the complete text-only SFT--GRPO pipeline than from text-only SFT under either audio-visual refinement method. With this initialization, audio-visual GRPO achieves the highest reasoning Macro Avg.\ among the staged configurations at 48.48 versus 47.42 for audio-visual SFT, with nearly identical perception averages of 78.50 and 78.53. This refinement uses about 90\% fewer training input tokens than full-data native audio-visual GRPO.

Compared with the complete text-only route in Tables~\ref{tab:rq1_main} and~\ref{tab:rq4_perception}, this configuration improves the perception Macro Avg.\ by 2.09 points while reducing reasoning by 0.91 points. It retains 93.5\% of the text-only reasoning gain measured by geometric mean improvement (24.15\% versus 25.83\% before refinement). It exceeds the full-data native audio-visual SFT--GRPO route by 1.11 reasoning Macro Avg.\ points, with a 0.57-point lower perception average. Together, these results establish our recommended text-centric post-training paradigm for Omni-LLMs: text-only SFT followed by GRPO for the main reasoning optimization, then reduced-data native audio-visual GRPO for perceptual refinement. This paradigm retains the main reasoning gains while restoring perception above the base level, addressing the performance--compute trade-off.

\section{Conclusion}
\label{sec:conclusion}

Building on our diagnostics, we studied text-centric post-training for Omni-LLMs.
Text-only training improves reasoning across data sources, model scales, and families, even with fully synthetic text requiring no audio-visual data in construction or training.
On Qwen2.5-Omni-7B, our best-performing text-only pipeline outperforms the complete native audio-visual route with less compute.
Reduced-data native audio-visual RL then restores perception while retaining most reasoning gains.
This text-centric paradigm organizes post-training around the capabilities to be improved: text for reasoning optimization and native audio-visual supervision for perceptual refinement.

\bibliography{iclr2027_conference}

\begin{thebibliography}{76}
\providecommand{\natexlab}[1]{#1}
\providecommand{\url}[1]{\texttt{#1}}
\expandafter\ifx\csname urlstyle\endcsname\relax
  \providecommand{\doi}[1]{doi: #1}\else
  \providecommand{\doi}{doi: \begingroup \urlstyle{rm}\Url}\fi

\bibitem[Akter et~al.(2026)Akter, Prabhumoye, Novikov, Han, Lin, Bakhturina, Nyberg, Choi, Patwary, Shoeybi, et~al.]{akter2026nemotron}
Syeda~Nahida Akter, Shrimai Prabhumoye, Matvei Novikov, Seungju Han, Ying Lin, Evelina Bakhturina, Eric Nyberg, Yejin Choi, Mostofa Patwary, Mohammad Shoeybi, et~al.
\newblock Nemotron-crossthink: Scaling self-learning beyond math reasoning.
\newblock In \emph{Proceedings of the 19th Conference of the European Chapter of the Association for Computational Linguistics (Volume 1: Long Papers)}, pp.\  984--1002, 2026.

\bibitem[Araujo et~al.(2026)Araujo, Bhati, Mirza, Kingsbury, Thomas, Feris, Glass, and Kuehne]{araujo2026avrt}
Edson Araujo, Saurabhchand Bhati, M~Jehanzeb Mirza, Brian Kingsbury, Samuel Thomas, Rogerio Feris, James~R Glass, and Hilde Kuehne.
\newblock Avrt: Audio-visual reasoning transfer through single-modality teachers.
\newblock \emph{arXiv preprint arXiv:2604.16617}, 2026.

\bibitem[Bousselham et~al.(2026)Bousselham, Kuehne, and Schmid]{bousselham2026vold}
Walid Bousselham, Hilde Kuehne, and Cordelia Schmid.
\newblock Vold: Reasoning transfer from llms to vision-language models via on-policy distillation.
\newblock In \emph{Proceedings of the IEEE/CVF Conference on Computer Vision and Pattern Recognition}, pp.\  26209--26218, 2026.

\bibitem[Cai et~al.(2026)Cai, Fu, Zhang, He, and Shan]{cai2026omnivideo}
Xinyue Cai, Chaoyou Fu, Yi-Fan Zhang, Ran He, and Caifeng Shan.
\newblock Omnivideo-100k: A dataset for audio-visual reasoning through structured scripts and evidence chains.
\newblock \emph{arXiv preprint arXiv:2606.14702}, 2026.

\bibitem[Cavalcanti et~al.(2024)Cavalcanti, Lessard, and Wilson]{cavalcanti2024adaptive}
Joao~V Cavalcanti, Laurent Lessard, and Ashia~C Wilson.
\newblock Adaptive backtracking line search.
\newblock \emph{arXiv preprint arXiv:2408.13150}, 2024.

\bibitem[Cawley \& Talbot(2010)Cawley and Talbot]{cawley2010overfitting}
Gavin~C Cawley and Nicola~LC Talbot.
\newblock On over-fitting in model selection and subsequent selection bias in performance evaluation.
\newblock \emph{The Journal of Machine Learning Research}, 11:\penalty0 2079--2107, 2010.

\bibitem[Chaudhry et~al.(2018)Chaudhry, Ranzato, Rohrbach, and Elhoseiny]{chaudhry2019agem}
Arslan Chaudhry, Marc'Aurelio Ranzato, Marcus Rohrbach, and Mohamed Elhoseiny.
\newblock Efficient lifelong learning with a-gem.
\newblock \emph{arXiv preprint arXiv:1812.00420}, 2018.

\bibitem[Chen et~al.(2025)Chen, Hu, Zhang, Chen, Wang, Li, Shyam, Zhou, Huang, Yang, et~al.]{chen2025omnixr}
Lichang Chen, Hexiang Hu, Mingda Zhang, Yiwen Chen, Zifeng Wang, Yandong Li, Pranav Shyam, Tianyi Zhou, Heng Huang, Ming-Hsuan Yang, et~al.
\newblock Omnixr: Evaluating omni-modality language models on reasoning across modalities.
\newblock In \emph{International Conference on Learning Representations}, volume 2025, pp.\  68100--68118, 2025.

\bibitem[Chen et~al.(2026{\natexlab{a}})Chen, Guo, Su, Li, Chen, Wu, Wang, Feng, Qu, and Cheng]{chen2026revisual}
Shuang Chen, Hangyu Guo, Zhaochen Su, Yafu Li, Jiacheng Chen, Yulun Wu, Weijie Wang, ZhiYuan Feng, Xiaoye Qu, and Yu~Cheng.
\newblock Revisual-r1: Advancing multimodal reasoning from optimized cold start to staged reinforcement learning.
\newblock In \emph{International Conference on Learning Representations}, volume 2026, pp.\  32869--32888, 2026{\natexlab{a}}.

\bibitem[Chen et~al.(2026{\natexlab{b}})Chen, Li, Xiong, Wang, Gao, Liu, Liu, Yang, Zeng, Zhang, et~al.]{chen2026omnireasoner}
Yu~Chen, Caorui Li, Ziyu Xiong, Yidong Wang, Mingqi Gao, Shuman Liu, Biao Liu, Chunfeng Yang, Anxiang Zeng, Haibo Zhang, et~al.
\newblock Omnireasoner: Thinking with long audio-video via native tool use.
\newblock \emph{arXiv preprint arXiv:2607.19339}, 2026{\natexlab{b}}.

\bibitem[Chen et~al.(2026{\natexlab{c}})Chen, Tao, Li, Hu, Chen, Yang, Yu, Jing, Zhang, Shao, et~al.]{chen2026omnivideo}
Zhangquan Chen, Jiale Tao, Ruihuang Li, Yihao Hu, Ruitao Chen, Zhantao Yang, Xinlei Yu, Haodong Jing, Manyuan Zhang, Shuai Shao, et~al.
\newblock Omnivideo-r1: Reinforcing audio-visual reasoning with query intention and modality attention.
\newblock \emph{arXiv preprint arXiv:2602.05847}, 2026{\natexlab{c}}.

\bibitem[Cheng et~al.(2026)Cheng, Li, Ma, Cai, Hu, Wang, Nie, and Li]{cheng2026omni}
Dongjie Cheng, Yongqi Li, Zhixin Ma, Hongru Cai, Yupeng Hu, Wenjie Wang, Liqiang Nie, and Wenjie Li.
\newblock Omni-r1: Towards the unified generative paradigm for multimodal reasoning.
\newblock In \emph{Findings of the Association for Computational Linguistics: ACL 2026}, pp.\  38907--38924, 2026.

\bibitem[Cheng et~al.(2025)Cheng, Ge, Wang, Ge, Liao, and Shan]{cheng2025video}
Junhao Cheng, Yuying Ge, Teng Wang, Yixiao Ge, Jing Liao, and Ying Shan.
\newblock Video-holmes: Can mllm think like holmes for complex video reasoning?
\newblock \emph{arXiv preprint arXiv:2505.21374}, 2025.

\bibitem[Dai et~al.(2026)Dai, Wu, Zeng, Hua, Liu, Li, Wang, Tong, Liang, Ma, et~al.]{dai2026latentomni}
Yifan Dai, Zhenhua Wu, Bohan Zeng, Daili Hua, Jialing Liu, Bozhou Li, Yuran Wang, Chengzhuo Tong, Hao Liang, Xiaochen Ma, et~al.
\newblock Latentomni: Rethinking omni-modal understanding via unified audio-visual latent reasoning.
\newblock \emph{arXiv preprint arXiv:2605.22012}, 2026.

\bibitem[Deshmukh et~al.(2026)Deshmukh, Chumachenko, Rintamaki, Le, Poon, Taheri, Karmanov, Liu, Seppanen, Goel, et~al.]{deshmukh2026nemotron}
Amala~Sanjay Deshmukh, Kateryna Chumachenko, Tuomas Rintamaki, Matthieu Le, Tyler Poon, Danial~Mohseni Taheri, Ilia Karmanov, Guilin Liu, Jarno Seppanen, Arushi Goel, et~al.
\newblock Nemotron 3 nano omni: efficient and open multimodal intelligence.
\newblock \emph{arXiv preprint arXiv:2604.24954}, 2026.

\bibitem[Du et~al.(2026)Du, Ye, Tang, Zhu, Lu, Cai, and Chen]{du2026openseeker}
Yuwen Du, Rui Ye, Shuo Tang, Xinyu Zhu, Yijun Lu, Yuzhu Cai, and Siheng Chen.
\newblock Openseeker: Democratizing frontier search agents by fully open-sourcing training data.
\newblock \emph{arXiv preprint arXiv:2603.15594}, 2026.

\bibitem[Face(2025)]{openr1}
Hugging Face.
\newblock Open r1: A fully open reproduction of deepseek-r1, January 2025.
\newblock URL \url{https://github.com/huggingface/open-r1}.

\bibitem[Farajtabar et~al.(2020)Farajtabar, Azizan, Mott, and Li]{farajtabar2020ogd}
Mehrdad Farajtabar, Navid Azizan, Alex Mott, and Ang Li.
\newblock Orthogonal gradient descent for continual learning.
\newblock In \emph{International conference on artificial intelligence and statistics}, pp.\  3762--3773. PMLR, 2020.

\bibitem[Farré et~al.(2024)Farré, Marafioti, Tunstall, Von~Werra, and Wolf]{Farré2024FineVideo}
Miquel Farré, Andi Marafioti, Lewis Tunstall, Leandro Von~Werra, and Thomas Wolf.
\newblock Finevideo.
\newblock \url{https://huggingface.co/datasets/HuggingFaceFV/finevideo}, 2024.

\bibitem[Fox \& Monette(2026)Fox and Monette]{fox2026cv}
John Fox and Georges Monette.
\newblock cv: An r package for cross-validating regression models.
\newblock \emph{Journal of Statistical Software}, 116:\penalty0 1--41, 2026.

\bibitem[Fu et~al.(2026)Fu, Yuan, Dong, Zhang, Shen, Hu, Li, Su, Long, Xie, et~al.]{fu2026video}
Chaoyou Fu, Haozhi Yuan, Yuhao Dong, Yi-Fan Zhang, Yunhang Shen, Xiaoxing Hu, Xueying Li, Jinsen Su, Chengwu Long, Xiaoyao Xie, et~al.
\newblock Video-mme-v2: Towards the next stage in benchmarks for comprehensive video understanding.
\newblock \emph{arXiv preprint arXiv:2604.05015}, 2026.

\bibitem[Guan et~al.(2026)Guan, Tu, Liang, Zhu, Ju, Luo, Luan, Liu, and Bai]{guan2026thinkomni}
Yiran Guan, Sifan Tu, Dingkang Liang, Linghao Zhu, Jianzhong Ju, Zhenbo Luo, Jian Luan, Yuliang Liu, and Xiang Bai.
\newblock Thinkomni: Lifting textual reasoning to omni-modal scenarios via guidance decoding.
\newblock \emph{arXiv preprint arXiv:2602.23306}, 2026.

\bibitem[Guha et~al.(2026)Guha, Marten, Keh, Raoof, Smyrnis, Bansal, Nezhurina, Mercat, Vu, Sprague, et~al.]{guha2026openthoughts}
Etash Guha, Ryan Marten, Sedrick Keh, Negin Raoof, Georgios Smyrnis, Hritik Bansal, Marianna Nezhurina, Jean Mercat, Trung Vu, Zayne Sprague, et~al.
\newblock Openthoughts: Data recipes for reasoning models.
\newblock In \emph{International Conference on Learning Representations}, volume 2026, pp.\  108059--108130, 2026.

\bibitem[Ho et~al.(2020)Ho, Nguyen, Sugawara, and Aizawa]{ho2020constructing}
Xanh Ho, Anh-Khoa~Duong Nguyen, Saku Sugawara, and Akiko Aizawa.
\newblock Constructing a multi-hop qa dataset for comprehensive evaluation of reasoning steps.
\newblock In \emph{Proceedings of the 28th International Conference on Computational Linguistics}, pp.\  6609--6625, 2020.

\bibitem[Hong et~al.(2026)Hong, Yan, Cai, Jiang, Hu, and Xie]{hong2026worldsense}
Jack Hong, Shilin Yan, Jiayin Cai, Xiaolong Jiang, Yao Hu, and Weidi Xie.
\newblock Worldsense: Evaluating real-world omnimodal understanding for multimodal llms.
\newblock In \emph{International Conference on Learning Representations}, volume 2026, pp.\  52423--52443, 2026.

\bibitem[Huang et~al.(2025)Huang, Tian, Maneechotesuwan, Chopra, and Kira]{huang2025digrap}
Chengyue Huang, Junjiao Tian, Brisa Maneechotesuwan, Shivang Chopra, and Zsolt Kira.
\newblock Directional gradient projection for robust fine-tuning of foundation models.
\newblock In \emph{International Conference on Learning Representations}, volume 2025, pp.\  60524--60544, 2025.

\bibitem[Jia et~al.(2026)Jia, Zhu, Zhong, Liu, Xi, Chen, Qin, Yang, Luo, and Shen]{jia2026omnijigsaw}
Yiduo Jia, Muzhi Zhu, Hao Zhong, Mingyu Liu, Yuling Xi, Hao Chen, Bin Qin, Yongjie Yang, Zhenbo Luo, and Chunhua Shen.
\newblock Omnijigsaw: Enhancing omni-modal reasoning via modality-orchestrated reordering.
\newblock \emph{arXiv preprint arXiv:2604.08209}, 2026.

\bibitem[Kim et~al.(2023)Kim, Joo, Kim, Jang, Ye, Shin, and Seo]{kim2023cot}
Seungone Kim, Se~Joo, Doyoung Kim, Joel Jang, Seonghyeon Ye, Jamin Shin, and Minjoon Seo.
\newblock The cot collection: Improving zero-shot and few-shot learning of language models via chain-of-thought fine-tuning.
\newblock In \emph{Proceedings of the 2023 Conference on Empirical Methods in Natural Language Processing}, pp.\  12685--12708, 2023.

\bibitem[Kulkarni \& Fazli(2026)Kulkarni and Fazli]{kulkarni2026avatar}
Yogesh Kulkarni and Pooyan Fazli.
\newblock Avatar: Reinforcement learning to see, hear, and reason over video.
\newblock In \emph{Proceedings of the IEEE/CVF Conference on Computer Vision and Pattern Recognition}, pp.\  7912--7922, 2026.

\bibitem[Kumar(2025)]{kumar2025cotscientificrag}
Abhinav Kumar.
\newblock {CoT-Scientific-RAG-Reasoning}.
\newblock Hugging Face dataset, December 2025.
\newblock URL \url{https://huggingface.co/datasets/abhinavdread/CoT-Scientific-RAG-Reasoning/tree/9a883557fd4b5b958c1c8b8b431ba08f6ae39c42}.
\newblock Revision 9a88355; accessed 2026-08-25.

\bibitem[Labs(2025)]{bespoke_stratos}
Bespoke Labs.
\newblock Bespoke-stratos: The unreasonable effectiveness of reasoning distillation.
\newblock https://www.bespokelabs.ai/blog/bespoke-stratos-the-unreasonable-effectiveness-of-reasoning-distillation, 2025.
\newblock Accessed: 2025-01-22.

\bibitem[Li et~al.(2026{\natexlab{a}})Li, Chen, Ji, Xu, Cui, Li, Zhang, Song, Zhang, He, et~al.]{li2026omnivideobench}
Caorui Li, Yu~Chen, Yiyan Ji, Jin Xu, Zhenyu Cui, Shihao Li, Yuanxing Zhang, Zhenghao Song, Dingling Zhang, Ying He, et~al.
\newblock Omnivideobench: Towards audio-visual understanding evaluation for omni mllms.
\newblock In \emph{International Conference on Learning Representations}, volume 2026, pp.\  138214--138236, 2026{\natexlab{a}}.

\bibitem[Li et~al.(2022)Li, Wei, Tian, Xu, Wen, and Hu]{li2022learning}
Guangyao Li, Yake Wei, Yapeng Tian, Chenliang Xu, Ji-Rong Wen, and Di~Hu.
\newblock Learning to answer questions in dynamic audio-visual scenarios.
\newblock In \emph{2022 IEEE/CVF Conference on Computer Vision and Pattern Recognition (CVPR)}, pp.\  19086--19096. IEEE, 2022.

\bibitem[Li et~al.(2025)Li, Liu, Yao, Zhang, An, Wang, Sun, Kong, and Liu]{li2025temporal}
Lei Li, Yuanxin Liu, Linli Yao, Peiyuan Zhang, Chenxin An, Lean Wang, Xu~Sun, Lingpeng Kong, and Qi~Liu.
\newblock Temporal reasoning transfer from text to video.
\newblock In \emph{International Conference on Learning Representations}, volume 2025, pp.\  74070--74102, 2025.

\bibitem[Li et~al.(2026{\natexlab{b}})Li, Jiao, Jin, Li, Wang, Wang, Dong, Jin, Wang, Lu, et~al.]{li2026omnigaia}
Xiaoxi Li, Wenxiang Jiao, Jiarui Jin, Haoxuan Li, Hao Wang, Shijian Wang, Guanting Dong, Jiajie Jin, Yinuo Wang, Yuan Lu, et~al.
\newblock Omnigaia: Towards native omni-modal ai agents.
\newblock \emph{arXiv preprint arXiv:2602.22897}, 2026{\natexlab{b}}.

\bibitem[Liu et~al.(2026{\natexlab{a}})Liu, Ma, Zhang, Zhang, Zhang, Yang, and Tian]{liu2026boosting}
Che Liu, Lichao Ma, Xiangyu~Tony Zhang, Yuxin Zhang, Haoyang Zhang, Xuerui Yang, and Fei Tian.
\newblock Boosting omni-modal language models: Staged post-training with visually debiased evaluation.
\newblock \emph{arXiv preprint arXiv:2605.12034}, 2026{\natexlab{a}}.

\bibitem[Liu et~al.(2026{\natexlab{b}})Liu, Wang, Liu, Zhu, Liao, Wang, and Wang]{liu2026anchornet}
Hongcheng Liu, Pingjie Wang, Heyang Liu, Zhiyuan Zhu, Yusheng Liao, Yanfeng Wang, and Yu~Wang.
\newblock Anchornet: Adaptive anchor token enhancement in video-grounded dialogue generation.
\newblock \emph{IEEE Journal of Selected Topics in Signal Processing}, 2026{\natexlab{b}}.

\bibitem[Liu et~al.(2026{\natexlab{c}})Liu, Wang, Chen, Wang, Zhu, Hou, Wang, and Wang]{liu-etal-2026-cross}
Hongcheng Liu, Yuhao Wang, Zhe Chen, Pingjie Wang, Zhiyuan Zhu, Yixuan Hou, Yanfeng Wang, and Yu~Wang.
\newblock Cross-modal coreference alignment: Enabling reliable information transfer in omni-{LLM}s.
\newblock In Maria Liakata, Viviane~P. Moreira, Jiajun Zhang, and David Jurgens (eds.), \emph{Proceedings of the 64th Annual Meeting of the {A}ssociation for {C}omputational {L}inguistics (Volume 1: Long Papers)}, pp.\  26430--26453, San Diego, California, United States, July 2026{\natexlab{c}}. Association for Computational Linguistics.
\newblock ISBN 979-8-89176-390-6.
\newblock \doi{10.18653/v1/2026.acl-long.1217}.
\newblock URL \url{https://aclanthology.org/2026.acl-long.1217/}.

\bibitem[Liu et~al.(2025{\natexlab{a}})Liu, Wu, Yu, Su, Mishra, Ramesh, Ranjan, Manem, Sun, Wang, et~al.]{liu2025instella}
Jiang Liu, Jialian Wu, Xiaodong Yu, Yusheng Su, Prakamya Mishra, Gowtham Ramesh, Sudhanshu Ranjan, Chaitanya Manem, Ximeng Sun, Ze~Wang, et~al.
\newblock Instella: Fully open language models with stellar performance.
\newblock \emph{arXiv preprint arXiv:2511.10628}, 2025{\natexlab{a}}.

\bibitem[Liu et~al.(2025{\natexlab{b}})Liu, Zhang, Qin, Ossowski, Gu, Jin, Kiblawi, Preston, Wei, Vozila, et~al.]{liu2025x}
Qianchu Liu, Sheng Zhang, Guanghui Qin, Timothy Ossowski, Yu~Gu, Ying Jin, Sid Kiblawi, Sam Preston, Mu~Wei, Paul Vozila, et~al.
\newblock X-reasoner: Towards generalizable reasoning across modalities and domains.
\newblock \emph{arXiv preprint arXiv:2505.03981}, 2025{\natexlab{b}}.

\bibitem[Liu et~al.(2024)Liu, Dong, and Zhang]{liu2024tackling}
Xiulong Liu, Zhikang Dong, and Peng Zhang.
\newblock Tackling data bias in music-avqa: Crafting a balanced dataset for unbiased question-answering.
\newblock In \emph{Proceedings of the IEEE/CVF Winter Conference on Applications of Computer Vision}, pp.\  4478--4487, 2024.

\bibitem[Lozhkov et~al.(2025)Lozhkov, Kydlíček, Allal, Penedo, Beeching, Gallouédec, Habib, Tunstall, and von Werra]{lozhkov2025openr1math220k}
Anton Lozhkov, Hynek Kydlíček, Loubna~Ben Allal, Guilherme Penedo, Edward Beeching, Quentin Gallouédec, Nathan Habib, Lewis Tunstall, and Leandro von Werra.
\newblock Openr1-math-220k.
\newblock \url{https://huggingface.co/datasets/open-r1/OpenR1-Math-220k}, 2025.

\bibitem[Lu et~al.(2025)Lu, Zhong, Liu, Fu, and Jia]{lu2025arpo}
Fanbin Lu, Zhisheng Zhong, Shu Liu, Chi-Wing Fu, and Jiaya Jia.
\newblock Arpo: End-to-end policy optimization for gui agents with experience replay.
\newblock \emph{arXiv preprint arXiv:2505.16282}, 2025.

\bibitem[Ma et~al.(2026)Ma, Xu, Xing, Chu, Wang, He, Xu, Heng, Yu, Lin, et~al.]{ma2026omni}
Ziyang Ma, Ruiyang Xu, Zhenghao Xing, Yunfei Chu, Yuxuan Wang, Jinzheng He, Jin Xu, Pheng-Ann Heng, Kai Yu, Junyang Lin, et~al.
\newblock Omni-captioner: Data pipeline, models, and benchmark for omni detailed perception.
\newblock In \emph{International Conference on Learning Representations}, volume 2026, pp.\  132277--132304, 2026.

\bibitem[Moshkov et~al.(2025)Moshkov, Hanley, Sorokin, Toshniwal, Henkel, Schifferer, Du, and Gitman]{moshkov2025aimo}
Ivan Moshkov, Darragh Hanley, Ivan Sorokin, Shubham Toshniwal, Christof Henkel, Benedikt Schifferer, Wei Du, and Igor Gitman.
\newblock Aimo-2 winning solution: Building state-of-the-art mathematical reasoning models with openmathreasoning dataset.
\newblock \emph{arXiv preprint arXiv:2504.16891}, 2025.

\bibitem[Mukherjee et~al.(2026)Mukherjee, Negi, Ganu, and Kumaraguru]{mukherjee2026c}
Swapnanil Mukherjee, Agyeya Negi, Tanuja Ganu, and Ponnurangam Kumaraguru.
\newblock C$^3$po: Evaluating cross-modal composition and counterfactual performance in omnimodal models.
\newblock \emph{arXiv preprint arXiv:2608.05381}, 2026.

\bibitem[Peng et~al.(2026)Peng, Zhang, Zhang, You, Liu, Zhu, Yang, Xu, Geng, and Yang]{peng2026lmm}
Yingzhe Peng, Gongrui Zhang, Miaosen Zhang, Zhiyuan You, Jie Liu, Qipeng Zhu, Kai Yang, Xingzhong Xu, Xin Geng, and Xu~Yang.
\newblock Lmm-r1: Empowering 3b lmms with strong reasoning abilities with two-stage rule-based rl.
\newblock \emph{Frontiers of Computer Science}, 2026.

\bibitem[Qin et~al.(2026)Qin, Zheng, Wang, Li, Yuan, Chen, and Wang]{qin2026humansense}
Zheng Qin, Ruobing Zheng, Yabing Wang, Tianqi Li, Yi~Yuan, Jingdong Chen, and Le~Wang.
\newblock Humansense: From multimodal perception to empathetic context-aware responses through reasoning mllms.
\newblock In \emph{Proceedings of the AAAI Conference on Artificial Intelligence}, volume~40, pp.\  24973--24981, 2026.

\bibitem[Qiu et~al.(2026)Qiu, Zhang, Qiao, Guan, Zhang, and Nie]{qiu2026splitlora}
Haomiao Qiu, Miao Zhang, Ziyue Qiao, Weili Guan, Min Zhang, and Liqiang Nie.
\newblock Splitlora: Balancing stability and plasticity in continual learning through gradient space splitting.
\newblock In \emph{International Conference on Learning Representations}, volume 2026, pp.\  48458--48480, 2026.

\bibitem[Rouditchenko et~al.(2025)Rouditchenko, Bhati, Araujo, Thomas, Kuehne, Feris, and Glass]{rouditchenko2025omni}
Andrew Rouditchenko, Saurabhchand Bhati, Edson Araujo, Samuel Thomas, Hilde Kuehne, Rogerio Feris, and James Glass.
\newblock Omni-r1: Do you really need audio to fine-tune your audio llm?
\newblock In \emph{2025 IEEE Automatic Speech Recognition and Understanding Workshop (ASRU)}, pp.\  1--7. IEEE, 2025.

\bibitem[Shao et~al.(2024)Shao, Wang, Zhu, Xu, Song, Bi, Zhang, Zhang, Li, Wu, et~al.]{shao2024deepseekmath}
Zhihong Shao, Peiyi Wang, Qihao Zhu, Runxin Xu, Junxiao Song, Xiao Bi, Haowei Zhang, Mingchuan Zhang, YK~Li, Yang Wu, et~al.
\newblock Deepseekmath: Pushing the limits of mathematical reasoning in open language models.
\newblock \emph{arXiv preprint arXiv:2402.03300}, 2024.

\bibitem[Sung-Bin et~al.(2025)Sung-Bin, Hyun-Bin, Jung-Mok, Senocak, Chung, and Oh]{sung2025avhbench}
Kim Sung-Bin, Oh~Hyun-Bin, Lee Jung-Mok, Arda Senocak, Joon~Son Chung, and Tae-Hyun Oh.
\newblock Avhbench: A cross-modal hallucination benchmark for audio-visual large language models.
\newblock In \emph{International Conference on Learning Representations}, volume 2025, pp.\  24244--24271, 2025.

\bibitem[Team et~al.(2026)Team, Abd, Aggarwal, Algayres, Andreev, Bachem, Ballantyne, Brick, C{\u{a}}rbune, Casbon, et~al.]{team2026gemma}
Gemma Team, Sherif~El Abd, Vaibhav Aggarwal, Robin Algayres, Alek Andreev, Olivier Bachem, Ian Ballantyne, Cormac Brick, Victor C{\u{a}}rbune, Michelle Casbon, et~al.
\newblock Gemma 4 technical report.
\newblock \emph{arXiv preprint arXiv:2607.02770}, 2026.

\bibitem[Team et~al.(2025)Team, Bai, Bing, Chen, Chen, Chen, Chen, Chen, Dai, Dong, et~al.]{team2025mirothinker}
MiroMind Team, Song Bai, Lidong Bing, Carson Chen, Guanzheng Chen, Yuntao Chen, Zhe Chen, Ziyi Chen, Jifeng Dai, Xuan Dong, et~al.
\newblock Mirothinker: Pushing the performance boundaries of open-source research agents via model, context, and interactive scaling.
\newblock \emph{arXiv preprint arXiv:2511.11793}, 2025.

\bibitem[Team(2026)]{team2026qwen3}
Qwen Team.
\newblock Qwen3.5-omni technical report.
\newblock \emph{arXiv preprint arXiv:2604.15804}, 2026.

\bibitem[Trivedi et~al.(2022)Trivedi, Balasubramanian, Khot, and Sabharwal]{trivedi2022musique}
Harsh Trivedi, Niranjan Balasubramanian, Tushar Khot, and Ashish Sabharwal.
\newblock musique: Multihop questions via single-hop question composition.
\newblock \emph{Transactions of the Association for Computational Linguistics}, 10:\penalty0 539--554, 2022.

\bibitem[Wang et~al.(2026{\natexlab{a}})Wang, Liu, Huang, Yu, Wang, Sun, Wu, Yuille, Barsoum, and Liu]{wang2026xmodbench}
Xingrui Wang, Jiang Liu, Chao Huang, Xiaodong Yu, Ze~Wang, Ximeng Sun, Jialian Wu, Alan Yuille, Emad Barsoum, and Zicheng Liu.
\newblock Xmodbench: Benchmarking cross-modal capabilities and consistency in omni-language models.
\newblock In \emph{International Conference on Learning Representations}, volume 2026, pp.\  19508--19521, 2026{\natexlab{a}}.

\bibitem[Wang et~al.(2026{\natexlab{b}})Wang, Zhang, Tu, Xu, Du, Liang, Wang, Li, Li, Fei, et~al.]{wang2026avi}
Yaoting Wang, Ziyi Zhang, Wenming Tu, Shaoxuan Xu, Wenjie Du, Cheng Liang, Weijun Wang, Yuanchao Li, Guangyao Li, Hao Fei, et~al.
\newblock Avi-bench: Toward human-like audio-visual intelligence of omni-mllms.
\newblock \emph{arXiv preprint arXiv:2606.07643}, 2026{\natexlab{b}}.

\bibitem[Wang et~al.(2026{\natexlab{c}})Wang, Hou, Javadov, Akhtar, and Sachan]{wang2026compose}
Yucheng Wang, Yifan Hou, Aydin Javadov, Mubashara Akhtar, and Mrinmaya Sachan.
\newblock Compose and fuse: Revisiting the foundational bottlenecks in multimodal reasoning.
\newblock In \emph{International Conference on Learning Representations}, volume 2026, pp.\  11272--11301, 2026{\natexlab{c}}.

\bibitem[Wu et~al.(2026)Wu, Chen, Tu, Tang, Shi, Liu, Lu, Xie, and Zhou]{wu2026seeing}
Juncheng Wu, Hardy Chen, Haoqin Tu, Xianfeng Tang, Freda Shi, Hui Liu, Hanqing Lu, Cihang Xie, and Yuyin Zhou.
\newblock From seeing to thinking: Decoupling perception and reasoning improves post-training of vision-language models.
\newblock \emph{arXiv preprint arXiv:2605.20177}, 2026.

\bibitem[Wu et~al.(2025)Wu, Shi, Wu, Zhang, Lin, Tang, and Luo]{wu2025concise}
Yifan Wu, Jingze Shi, Bingheng Wu, Jiayi Zhang, Xiaotian Lin, Nan Tang, and Yuyu Luo.
\newblock Concise reasoning, big gains: Pruning long reasoning trace with difficulty-aware prompting.
\newblock \emph{arXiv preprint arXiv:2505.19716}, 2025.

\bibitem[Xiao et~al.(2021)Xiao, Shang, Yao, and Chua]{xiao2021next}
Junbin Xiao, Xindi Shang, Angela Yao, and Tat-Seng Chua.
\newblock Next-qa: Next phase of question-answering to explaining temporal actions.
\newblock In \emph{2021 IEEE/CVF Conference on Computer Vision and Pattern Recognition (CVPR)}, pp.\  9772--9781. IEEE, 2021.

\bibitem[Xie et~al.(2026)Xie, Huang, Ma, Luo, Yang, Ma, Chen, Zeng, Zou, Lu, et~al.]{xie2026socialomni}
Tianyu Xie, Jinfa Huang, Yuexiao Ma, Rongfang Luo, Yan Yang, Qingchuan Ma, Wang Chen, Yuhui Zeng, Yixuan Zou, Zhiqiang Lu, et~al.
\newblock Socialomni: Benchmarking audio-visual social interactivity in omni models.
\newblock \emph{arXiv preprint arXiv:2603.16859}, 2026.

\bibitem[Xing et~al.(2025)Xing, Hu, Fu, Wang, Dai, and Heng]{xing2025echoink}
Zhenghao Xing, Xiaowei Hu, Chi-Wing Fu, Wenhai Wang, Jifeng Dai, and Pheng-Ann Heng.
\newblock Echoink-r1: Exploring audio-visual reasoning in multimodal llms via reinforcement learning.
\newblock \emph{arXiv preprint arXiv:2505.04623}, 2025.

\bibitem[Xu et~al.(2025{\natexlab{a}})Xu, Guo, He, Hu, He, Bai, Chen, Wang, Fan, Dang, Zhang, Wang, Chu, and Lin]{Qwen2.5-Omni}
Jin Xu, Zhifang Guo, Jinzheng He, Hangrui Hu, Ting He, Shuai Bai, Keqin Chen, Jialin Wang, Yang Fan, Kai Dang, Bin Zhang, Xiong Wang, Yunfei Chu, and Junyang Lin.
\newblock Qwen2.5-omni technical report.
\newblock \emph{arXiv preprint arXiv:2503.20215}, 2025{\natexlab{a}}.

\bibitem[Xu et~al.(2025{\natexlab{b}})Xu, Guo, Hu, Chu, Wang, He, Wang, Shi, He, Zhu, et~al.]{xu2025qwen3}
Jin Xu, Zhifang Guo, Hangrui Hu, Yunfei Chu, Xiong Wang, Jinzheng He, Yuxuan Wang, Xian Shi, Ting He, Xinfa Zhu, et~al.
\newblock Qwen3-omni technical report.
\newblock \emph{arXiv preprint arXiv:2509.17765}, 2025{\natexlab{b}}.

\bibitem[Xu et~al.(2026)Xu, Wang, Cheng, Liu, Wang, and Wang]{xu2026agentic}
Ke~Xu, Yuhao Wang, Ziyang Cheng, Hongcheng Liu, Yanfeng Wang, and Yu~Wang.
\newblock Agentic active omni-modal perception for multi-hop audio-visual reasoning.
\newblock \emph{arXiv preprint arXiv:2605.28192}, 2026.

\bibitem[Yang et~al.(2025{\natexlab{a}})Yang, Le, Xing, An, Chen, Zhao, Song, and Zhang]{yang2025toolmind}
Chen Yang, Ran Le, Yun Xing, Zhenwei An, Zongchao Chen, Wayne~Xin Zhao, Yang Song, and Tao Zhang.
\newblock Toolmind technical report: A large-scale, reasoning-enhanced tool-use dataset.
\newblock \emph{arXiv preprint arXiv:2511.15718}, 2025{\natexlab{a}}.

\bibitem[Yang et~al.(2022)Yang, Wang, Duan, Chen, Hou, Jin, and Zhu]{yang2022avqa}
Pinci Yang, Xin Wang, Xuguang Duan, Hong Chen, Runze Hou, Cong Jin, and Wenwu Zhu.
\newblock Avqa: A dataset for audio-visual question answering on videos.
\newblock In \emph{Proceedings of the 30th ACM international conference on multimedia}, pp.\  3480--3491, 2022.

\bibitem[Yang et~al.(2025{\natexlab{b}})Yang, Yao, Chen, Fu, Bai, Zhao, Sun, Yin, Wei, and Zhou]{yang2025humanomniv2}
Qize Yang, Shimin Yao, Weixuan Chen, Shenghao Fu, Detao Bai, Jiaxing Zhao, Boyuan Sun, Bowen Yin, Xihan Wei, and Jingren Zhou.
\newblock Humanomniv2: From understanding to omni-modal reasoning with context.
\newblock \emph{arXiv preprint arXiv:2506.21277}, 2025{\natexlab{b}}.

\bibitem[Yang et~al.(2018)Yang, Qi, Zhang, Bengio, Cohen, Salakhutdinov, and Manning]{yang2018hotpotqa}
Zhilin Yang, Peng Qi, Saizheng Zhang, Yoshua Bengio, William Cohen, Ruslan Salakhutdinov, and Christopher~D Manning.
\newblock Hotpotqa: A dataset for diverse, explainable multi-hop question answering.
\newblock In \emph{Proceedings of the 2018 conference on empirical methods in natural language processing}, pp.\  2369--2380, 2018.

\bibitem[Yao et~al.(2026)Yao, Zhang, Huang, Long, Zhang, Zou, Xiong, Rong, Wu, Su, et~al.]{yao2026lens}
Ruilin Yao, Bo~Zhang, Jirui Huang, Xinwei Long, Yifang Zhang, Tianyu Zou, Shili Xiong, Yi~Rong, Yufei Wu, Shichao Su, et~al.
\newblock Lens: Multi-level evaluation of multimodal reasoning with large language models.
\newblock In \emph{International Conference on Learning Representations}, volume 2026, pp.\  2627--2654, 2026.

\bibitem[Yu et~al.(2025)Yu, Zhang, Qiao, Qin, and Xiong]{yu2025unicorn}
Xiaomin Yu, Wenjie Zhang, Ziyue Qiao, Chengwei Qin, and Hui Xiong.
\newblock Text-only data synthesis for vision language model training.
\newblock \emph{arXiv preprint arXiv:2503.22655}, 2025.

\bibitem[Zhang et~al.(2025)Zhang, Liu, Li, Chen, and You]{zhang2025seeingeye}
Weijia Zhang, Zijia Liu, Haoru Li, Haoqi Chen, and Jiaxuan You.
\newblock Seeingeye: Agentic information flow unlocks multimodal reasoning in text-only llms.
\newblock \emph{arXiv preprint arXiv:2510.25092}, 2025.

\bibitem[Zhong et~al.(2025)Zhong, Zhu, Du, Huang, Zhao, Liu, Wang, Chen, and Shen]{zhong2026omni}
Hao Zhong, Muzhi Zhu, Zongze Du, Zheng Huang, Canyu Zhao, Mingyu Liu, Wen Wang, Hao Chen, and Chunhua Shen.
\newblock Omni-r1: Reinforcement learning for omnimodal reasoning via two-system collaboration.
\newblock In D.~Belgrave, C.~Zhang, H.~Lin, R.~Pascanu, P.~Koniusz, M.~Ghassemi, and N.~Chen (eds.), \emph{Advances in Neural Information Processing Systems}, volume 38, Main Conference, pp.\  45266--45297. Curran Associates, Inc., 2025.
\newblock \doi{10.52202/085713-1508}.
\newblock URL \url{https://proceedings.neurips.cc/paper_files/paper/2025/file/4086fe59dc3584708468fba0e459f6a7-Paper-Conference.pdf}.

\bibitem[Zhou et~al.(2025)Zhou, Wang, Wu, and Jiang]{zhou2025daily}
Ziwei Zhou, Rui Wang, Zuxuan Wu, and Yu-Gang Jiang.
\newblock Daily-omni: Towards audio-visual reasoning with temporal alignment across modalities.
\newblock \emph{arXiv preprint arXiv:2505.17862}, 2025.

\end{thebibliography}
\bibliographystyle{iclr2027_conference}

\newpage
\appendix

\section{Diagnostic Analysis}
\label{sec:appendix_diagnostic}

\subsection{Diagnostic Question Construction and Evaluation}
\label{sec:appendix_diagnostic_questions}

This section details the construction and evaluation of the diagnostic questions used in Section~\ref{sec:preliminary} to examine perception and reasoning in Omni-LLMs.

\paragraph{Question taxonomy.}
We categorize questions by the number of reasoning steps specified during construction. Single-hop perception questions assess the direct extraction of information from audio-visual inputs. Multi-hop questions combine facts from corresponding single-hop QA pairs across audio and vision in two or three steps. They must connect entities, events, attributes, or relations rather than merely group independent questions from the same video or topic. Both question types cover five domains: Causal, Character, Modality, Temporal, and Theme.

\paragraph{Data construction.}
We use FineVideo's audio-visual data and annotations \citep{Farré2024FineVideo} to generate QA pairs with Qwen3.7-Max and convert them into four-option MCQs. All nine models independently answer the original set of approximately 4,000 single-hop questions, which remains the full 1-hop evaluation set in Section~\ref{sec:preliminary}. Only questions answered correctly by all nine models are used to construct multi-hop questions. Each multi-hop question combines facts from these single-hop QA pairs, supported by relevant audio or visual segments. Together with the question and stated task rules, these facts must uniquely determine the answer without unstated specialized knowledge. We remove ambiguous, underspecified, or unsupported questions, as well as cases solvable from a single cue or commonsense even with answer options. Together, the original single-hop and resulting multi-hop sets draw on 2,703 videos; the latter contains approximately 5,000 questions, equally divided between 2-hop and 3-hop.

\paragraph{Evaluation protocol.}
Table~\ref{tab:diagnostic_accuracy} lists the evaluated models and their accuracy at each hop count. Models receive the question and its four options with the corresponding audio-visual inputs. For each model, a multi-hop question and its corresponding single-hop questions use the same audio-visual material and preprocessing settings, with each question answered in a separate call. Multi-hop evaluation provides no preceding single-hop questions, answers, evidence hints, or conversation history. All models use the original single-hop set and the final multi-hop sets defined above. Accuracy is the fraction of correctly answered questions within each set. For Figure~\ref{fig:hop_gap}, domain scores are averaged across models, and the macro-average assigns equal weight to all five domains.

\begin{table}[htbp]
\centering
\caption{Accuracy (\%) by model and question hop count in the diagnostic study.}
\label{tab:diagnostic_accuracy}
\fontsize{8}{12}\selectfont
\renewcommand{\arraystretch}{1.2}
\setlength{\aboverulesep}{1.5pt}
\setlength{\belowrulesep}{1.5pt}
\begin{threeparttable}
\begin{tabular*}{\linewidth}{@{\extracolsep{\fill}}lrrr@{}}
\toprule
\textbf{Model} & \textbf{1-hop} & \textbf{2-hop} & \textbf{3-hop} \\
\midrule
AVATAR-7B \citep{kulkarni2026avatar} & 93.4 & 62.0 & 60.5 \\

HumanSense \citep{qin2026humansense} & 94.6 & 64.6 & 65.2 \\

Omni-R1 \citep{zhong2026omni} & 94.0 & 65.0 & 62.5 \\

EchoInk-R1-7B \citep{xing2025echoink} & 95.5 & 71.0 & 68.4 \\

HumanOmniV2-7B \citep{yang2025humanomniv2} & 96.3 & 71.8 & 72.2 \\

OmniVideo-7B \citep{cai2026omnivideo} & 95.1 & 69.5 & 67.5 \\

Qwen2.5-Omni-3B \citep{Qwen2.5-Omni} & 90.9 & 37.4 & 30.0 \\

Qwen2.5-Omni-7B \citep{Qwen2.5-Omni} & 93.8 & 60.0 & 58.0 \\

Qwen3-Omni-30B-A3B-Instruct \citep{xu2025qwen3} & 98.5 & 75.2 & 76.0 \\
\bottomrule
\end{tabular*}
\end{threeparttable}
\end{table}

\subsection{Local Optimization Analysis}
\label{sec:appendix_gradient_analysis}

\paragraph{Experimental setup.}
We use Qwen2.5-Omni-3B with the full question sets defined above. Each objective averages the cross-entropy loss for predicting the correct answer token, using probabilities normalized over the full vocabulary. We compute gradients for the Thinker's Transformer decoder layers and final normalization layer, with all other parameters fixed. Gradients are averaged across questions before normalization.

We partition the data into five video-level folds, keeping all questions from each video in the same fold \citep{fox2026cv}. For each split, gradient estimation for constructing update directions and update-norm calibration use only the four training folds. All updates start from the same initial checkpoint and are evaluated on the held-out fold, which does not inform update construction or step-size selection \citep{cawley2010overfitting}.

\paragraph{First-order analysis.}
Gradient projection is used to constrain parameter updates in multi-objective fine-tuning and continual learning \citep{farajtabar2020ogd,chaudhry2019agem,huang2025digrap,qiu2026splitlora}. Within each split, the following analysis uses the training-fold objectives and gradients, with the split index omitted for clarity. We write an update as $-\eta u$, where $u=g_t/\|g_t\|_2$ for the original update and $u=r_{t\perp s}/\|r_{t\perp s}\|_2$ for the projected update in Equation~\ref{eq:local_projection}. Both directions have unit norm, so $\eta$ is the Euclidean norm of the parameter change. The first-order change in the other training objective is $-\eta g_s^\top u$; Equation~\ref{eq:local_projection} sets this term to zero. For $c=g_E^\top g_R/(\|g_E\|_2\|g_R\|_2)$, the maximum first-order decrease in the target loss per unit update norm satisfies
\begin{equation}
\max_{\|u\|_2\le1,\;g_s^\top u=0}g_t^\top u
=\|r_{t\perp s}\|_2,
\qquad
\frac{\|r_{t\perp s}\|_2}{\|g_t\|_2}=\sqrt{1-c^2}.
\end{equation}
Here $u$ ranges over feasible directions. For any such $u$, $g_t^\top u=r_{t\perp s}^\top u$; Cauchy--Schwarz gives the bound. When $r_{t\perp s}\ne0$, equality is attained at $u=r_{t\perp s}/\|r_{t\perp s}\|_2$, whose unit norm matches the update in Equation~\ref{eq:local_projection}. Across the five splits, the mean gradient cosine similarity is 0.5391, and the mean of the split-specific first-order retention ratios is 84.23\% at the same update norm. Held-out loss responses are measured after applying each split-specific update; training-gradient orthogonality does not impose a zero first-order change on held-out objectives.

\begin{wraptable}[13]{r}{0.5\textwidth}
\vspace{\dimexpr-\intextsep-1.4pt\relax}
\centering
\begin{minipage}{\linewidth}
\caption{First-order retention (\%) on held-out objectives, averaged over three seeds. Values are projected-to-original ratios of predicted loss decrease.}
\label{tab:diagnostic_control_estimates}
\fontsize{8}{9.5}\selectfont
\setlength{\tabcolsep}{2.5pt}
\renewcommand{\arraystretch}{1.04}
\setlength{\aboverulesep}{1.5pt}
\setlength{\belowrulesep}{1.5pt}
\begin{tabularx}{\linewidth}{@{}Xr@{\hspace{8pt}}r@{}}
\toprule
\raisebox{\dimexpr0.5\baselineskip+1.5pt\relax}{\textbf{Update}} & \shortstack[r]{\strut\textbf{Target}\\\strut\textbf{retention}} & \shortstack[r]{\strut\textbf{Other}\\\strut\textbf{retention}} \\
\midrule
$E_1$ projected ($\perp E_2$) & 56.6 & 53.6 \\
$E_2$ projected ($\perp E_1$) & 60.9 & 58.5 \\
\midrule
$E_1$ projected ($\perp R$) & 71.8 & 5.7 \\
$R$ projected ($\perp E_1$) & 85.0 & 13.1 \\
\bottomrule
\end{tabularx}
\end{minipage}
\end{wraptable}

We compare random perception halves $E_1/E_2$ with a size-matched $E_1/R$ pair, projecting in both directions. The full reasoning set contains equal numbers of 2-hop and 3-hop questions. Sampling is stratified by hop count and question type, with independent held-out perception splits. We pool held-out directional derivatives across folds by question count before computing projected-to-original ratios. \mbox{Table~\ref{tab:diagnostic_control_estimates}} shows similar retention on the two perception halves, whereas cross-task projections show higher target retention and lower retention on the other objective.

\noindent\begin{minipage}{\linewidth}
\captionsetup{type=table}

\centering

\caption{Loss responses at two update norms, with $R$ also reported separately for 2-hop and 3-hop questions. The higher-step $E$/$R$ responses correspond to Figure~\ref{fig:local_optimization}. All values are percentages; positive reductions indicate loss decreases.}

\label{tab:diagnostic_responses}

\fontsize{8}{12}\selectfont
\renewcommand{\arraystretch}{1.2}
\setlength{\aboverulesep}{1.5pt}
\setlength{\belowrulesep}{1.5pt}
\setlength{\tabcolsep}{4pt}

\begin{tabular*}{\linewidth}{@{\extracolsep{\fill}}llrrrrr@{}}

\toprule

\textbf{Step} & \textbf{Update} & \textbf{\boldmath $E$ reduction} & \multicolumn{3}{c}{\textbf{\boldmath $R$ reduction}} & \textbf{Target retention} \\

\cmidrule(lr){4-6}

& & & \textbf{Overall} & \textbf{2-hop} & \textbf{3-hop} & \\

\midrule

Lower & E original & +1.675 & +1.910 & +1.867 & +1.985 & 100.00 \\

Lower & E projected & +1.130 & +0.006 & +0.008 & +0.003 & 67.48 \\

Lower & R original & +1.176 & +3.331 & +2.718 & +4.397 & 100.00 \\

Lower & R projected & -0.040 & +2.860 & +2.077 & +4.224 & 85.89 \\

\midrule

Higher & E original & +3.632 & +4.333 & +4.201 & +4.563 & 100.00 \\

Higher & E projected & +2.699 & +0.362 & +0.466 & +0.181 & 74.30 \\

Higher & R original & +2.080 & +7.358 & +5.694 & +10.254 & 100.00 \\

Higher & R projected & +0.060 & +6.013 & +4.150 & +9.255 & 81.72 \\

\bottomrule

\end{tabular*}

\end{minipage}

\begin{table}[!t]

\centering

\caption{Loss reductions (\%) by held-out video-level fold at both update norms.}

\label{tab:diagnostic_exploratory}

\fontsize{8}{12}\selectfont
\renewcommand{\arraystretch}{1.5}
\setlength{\aboverulesep}{1.5pt}
\setlength{\belowrulesep}{1.5pt}

\setlength{\tabcolsep}{4pt}

\begin{tabular*}{\linewidth}{@{\extracolsep{\fill}}crrrrrrrr@{}}

\toprule

& \multicolumn{2}{c}{\textbf{\boldmath $E$ original}} & \multicolumn{2}{c}{\textbf{\boldmath $E$ projected}} & \multicolumn{2}{c}{\textbf{\boldmath $R$ original}} & \multicolumn{2}{c}{\textbf{\boldmath $R$ projected}} \\

\cmidrule(lr){2-3}\cmidrule(lr){4-5}\cmidrule(lr){6-7}\cmidrule(lr){8-9}

\textbf{Step / held-out fold} & \multicolumn{1}{c}{\boldmath $L_E$} & \multicolumn{1}{c}{\boldmath $L_R$} & \multicolumn{1}{c}{\boldmath $L_E$} & \multicolumn{1}{c}{\boldmath $L_R$} & \multicolumn{1}{c}{\boldmath $L_E$} & \multicolumn{1}{c}{\boldmath $L_R$} & \multicolumn{1}{c}{\boldmath $L_E$} & \multicolumn{1}{c}{\boldmath $L_R$} \\

\midrule

Lower / 1 & +1.181 & +1.944 & +0.934 & +0.006 & +0.834 & +3.106 & -0.042 & +2.272 \\

Lower / 2 & +2.441 & +1.589 & +1.407 & +0.006 & +1.417 & +3.252 & -0.032 & +2.854 \\

Lower / 3 & +1.474 & +1.678 & +0.956 & +0.006 & +1.183 & +3.416 & -0.060 & +2.902 \\

Lower / 4 & +1.555 & +1.850 & +1.272 & -0.002 & +0.887 & +2.789 & -0.048 & +2.655 \\

Lower / 5 & +1.749 & +2.439 & +1.132 & +0.015 & +1.490 & +4.020 & -0.020 & +3.524 \\

\midrule

Lower / Pooled & +1.675 & +1.910 & +1.130 & +0.006 & +1.176 & +3.331 & -0.040 & +2.860 \\

\midrule

Higher / 1 & +2.794 & +3.986 & +2.342 & +0.296 & +1.739 & +7.247 & +0.076 & +6.037 \\

Higher / 2 & +4.248 & +4.072 & +3.289 & +0.354 & +2.669 & +7.180 & +0.068 & +5.959 \\

Higher / 3 & +3.748 & +4.299 & +2.877 & +0.385 & +1.848 & +7.533 & +0.031 & +5.952 \\

Higher / 4 & +3.720 & +4.505 & +2.401 & +0.331 & +1.961 & +6.991 & +0.039 & +5.529 \\

Higher / 5 & +3.628 & +4.734 & +2.544 & +0.435 & +2.208 & +7.794 & +0.091 & +6.546 \\

\midrule

Higher / Pooled & +3.632 & +4.333 & +2.699 & +0.362 & +2.080 & +7.358 & +0.060 & +6.013 \\

\bottomrule

\end{tabular*}

\end{table}

\paragraph{Step sizes.}

Within each split, we calibrate the higher update norm on 128 questions drawn from the four training folds: 64 perception questions and 64 reasoning questions, equally divided between 2-hop and 3-hop. Starting from an initial candidate norm of approximately $1.44\times10^{-2}$, we halve the candidate norm until both original update directions satisfy the Armijo condition on their respective calibration objectives, using the calibration losses and directional derivatives with $\sigma=0.1$ \citep{cavalcanti2024adaptive}. This search selects $\eta_{\rm high}\approx7.20\times10^{-3}$ in all five splits. We then set $\eta_{\rm low}=\eta_{\rm high}/2\approx3.60\times10^{-3}$ to assess sensitivity to the update norm. In each split, we verify that this lower norm also satisfies the Armijo condition for both original directions on their respective calibration objectives. All four update directions use both norms, with each update starting from $\theta_0$.

After fixing both norms, we check sufficient decrease on each held-out fold for all four split-specific update directions at both norms, without modifying either norm. Let $L_t^{\rm val}$ denote the held-out objective and $h_t=\nabla L_t^{\rm val}(\theta_0)$. For $D_t=L_t^{\rm val}(\theta_0)-L_t^{\rm val}(\theta_0-\eta u)$, the Armijo condition is $D_t\ge\sigma\eta h_t^\top u$, with $\sigma=0.1$ and $h_t^\top u>0$. Across the 40 checks (eight per split), $D_t/(\eta h_t^\top u)$ ranges from 0.328 to 0.512, satisfying the Armijo condition in every case. This establishes sufficient decrease on the held-out objectives without requiring accurate first-order predictions of the loss reduction.

\paragraph{Measured responses.}
For cross-validation, each question is evaluated only under the update constructed without its video-level fold. We pool question-level baseline losses and loss reductions across the five held-out folds before computing percentages, rather than averaging fold-level percentages. Writing the pooled mean loss reduction as $\bar D_t$ and the pooled initial mean loss as $\bar L_t(\theta_0)$, we report relative loss reduction as $100\bar D_t/\bar L_t(\theta_0)$ and target retention as $100\bar D_t^{\rm projected}/\bar D_t^{\rm original}$. For $R$, losses are pooled across 2-hop and 3-hop questions before computing percentages. Table~\ref{tab:diagnostic_responses} reports both update norms; its higher-step overall responses match Figure~\ref{fig:local_optimization}. At the lower step, the projected perception update retains 67.48\% of its target loss reduction and decreases reasoning loss by 0.006\%. The projected reasoning update retains 85.89\% of its target loss reduction and increases perception loss by 0.040\%. Thus, the lower step also supports local partial decoupling of the two mean-loss objectives. Table~\ref{tab:diagnostic_exploratory} gives the held-out-fold breakdown.

\section{Training and Evaluation Details}
\label{sec:appendix_experimental_details}

\subsection{Training Setup}
\label{sec:appendix_training_setup}

\paragraph{Training data and preprocessing.}
Table~\ref{tab:training_data} lists the training data types and sources. Text-only and audio-visual GRPO use the same questions in the main comparison. For RQ3, Qwen3.7-Max generates fictional scripts and MCQs with answers and reasoning explanations from abstract scenarios, task types, and aggregate distribution statistics, without native audio-visual inputs (Figure~\ref{fig:rq3_synthesis_pipeline}). Scenarios cover daily life, work and craft, entertainment, sports, and social interaction. Programmatic checks validate structure, timing, and answer formats; model-based checks assess consistency, answer uniqueness, and distractor plausibility. We manually review a subset of synthetic samples. Generation and validation prompts appear in Appendix~\ref{sec:appendix_synthetic_prompts}.

\paragraph{Training settings.}
We apply LoRA ($r=16$, $\alpha=32$) to the Thinker's Transformer decoder in Qwen-Omni and the language model in Gemma-4. Audio and vision encoders and multimodal projection modules remain frozen. SFT and GRPO use AdamW with linear learning-rate warmup over the first 5\% of training steps, followed by cosine decay. Table~\ref{tab:appendix_training_hyperparameters} lists the hyperparameters.

GRPO clips probability ratios to $[0.8,1.2]$ ($\epsilon=0.2$). Qwen2.5-Omni-7B uses the same SFT and GRPO hyperparameters for text-only and audio-visual training. For audio-visual training, we retain audio and sample frames at 1 FPS, with up to 180 frames per video and 200,704 pixels per frame. Other settings, including sequence-length limits, follow official base-model defaults.

\begin{table}[htbp]
\centering
\caption{SFT and GRPO settings. GRPO batch sizes count completions; ``--'' marks unused stages.}
\label{tab:appendix_training_hyperparameters}
\scriptsize
\setlength{\tabcolsep}{1pt}
\renewcommand{\arraystretch}{1.2}
\begin{tabular*}{\linewidth}{@{\extracolsep{\fill}}l*{5}{c}@{}}
\toprule
\textbf{Hyperparameter} & \textbf{Qwen2.5-Omni-7B} & \textbf{Qwen2.5-Omni-3B} & \textbf{Gemma-4-E4B-IT} & \textbf{Gemma-4-E2B-IT} & \textbf{Qwen3-Omni-30B-A3B-Instruct} \\
\midrule
\multicolumn{6}{@{}l}{\textbf{SFT}} \\
Learning rate & $2\times10^{-6}$ & $5\times10^{-6}$ & $5\times10^{-5}$ & $5\times10^{-5}$ & $10^{-4}$ \\
Global batch & $48$ & $48$ & $48$ & $48$ & $8$ \\
Epochs & $1$ & $3$ & $1$ & $1$ & $1$ \\
\midrule
\multicolumn{6}{@{}l}{\textbf{GRPO}} \\
Learning rate & $10^{-5}$ & $10^{-5}$ & $10^{-5}$ & $10^{-5}$ & -- \\
Global batch & $16$ & $16$ & $16$ & $16$ & -- \\
Epochs & $1$ & $1$ & $1$ & $1$ & -- \\
Generations per prompt & $8$ & $8$ & $8$ & $8$ & -- \\
Temperature & $1.0$ & $1.0$ & $1.0$ & $1.0$ & -- \\
Top-$p$ & $0.99$ & $0.99$ & $0.95$ & $0.95$ & -- \\
Top-$k$ & $50$ & $50$ & $64$ & $64$ & -- \\
KL coefficient $\beta$ & $0.001$ & $0.001$ & $0.001$ & $0.001$ & -- \\
Clipping threshold $\epsilon$ & $0.2$ & $0.2$ & $0.2$ & $0.2$ & -- \\
Repetition penalty & $1.0$ & $1.0$ & $1.0$ & $1.0$ & -- \\
\bottomrule
\end{tabular*}
\end{table}

\paragraph{Rewards.}
GRPO combines rule-based accuracy and format rewards as $R=R_{\mathrm{acc}}+0.1R_{\mathrm{format}}$. The accuracy reward checks extracted final answers against references using task-specific verification, including option matching for MCQs and answer verification for mathematical problems. The format reward checks the required structure for reasoning and final answers.

\paragraph{Implementation and compute accounting.}
We train on eight NVIDIA A100 PCIe 80\,GB GPUs. We reproduce external baselines with their official training datasets and original training stages, measuring GPU-hours on the same hardware. We count non-padding training input tokens using each model's tokenizer and training template, including special tokens and expanded audio and visual token positions. $\mathcal{I}$ indexes sample occurrences across all epochs within each stage. $L_i^{\mathrm{SFT}}$~denotes the complete sequence length including assistant targets, and $L_i^{\mathrm{prompt}}$ denotes the prompt length before generation. With $G=8$ completions per prompt, counts in billions are
\begin{equation}
\label{eq:training_input_tokens}
T_{\mathrm{SFT}}=10^{-9}\sum_{i\in\mathcal{I}}L_i^{\mathrm{SFT}},
\qquad
T_{\mathrm{GRPO}}=10^{-9}G\sum_{i\in\mathcal{I}}L_i^{\mathrm{prompt}}.
\end{equation}
$T_{\mathrm{GRPO}}$ excludes generated completion tokens.

Training compute is $H_{\mathrm{GPU}}=\sum_j W_jt_j/3600$ GPU-hours, with $W_j$ GPUs and wall-clock duration $t_j$ (seconds) per interval $j$. GRPO includes online rollout generation and policy optimization.

\noindent\begin{minipage}{\linewidth}
\subsection{Evaluation Setup}
\label{sec:appendix_evaluation_setup}

We follow each benchmark's official evaluation settings for decoding, maximum output length, and audio-visual sampling. For Gemma-4, we retain the benchmark protocols but use official model inference defaults and set the maximum output length to 1,024 tokens. Qwen3.7-Max assesses response quality on SocialOmni How and answers Omni-Cloze's cloze-style MCQs using captions generated by the evaluated model. For Video-MME-v2, we report average accuracy (Avg Acc). For perceptual evaluation, we restrict AVQA to the Which, Come From, Happening, and Where question types, and NExT-QA to the Descriptive subset. AVHBench Overall is the pooled accuracy across Audio-driven Video Hallucination and Video-driven Audio Hallucination. Macro Avg.\ is the unweighted mean of nine reasoning scores or five perception scores: AVHBench Overall, MUSIC-AVQA v1 and v2, AVQA, and NExT-QA. Geometric Mean Improvement uses the same score sets. Evaluation system prompts appear in Appendix~\ref{sec:appendix_training_evaluation_prompts}.
\end{minipage}

\section{Further Analyses}
\label{sec:appendix_further_analyses}
\subsection{Training Data Mixtures}
To complement RQ2, we mix Script-based MCQs with either open-ended questions or General reasoning data during SFT, using CoT supervision. The subsequent GRPO phase uses the same Script-based MCQ data for all configurations. Table~\ref{tab:appendix_rq2_data} reports results for these mixtures, while Table~\ref{tab:rq2_data} provides the reference results without data mixing.

\label{sec:appendix_training_data}

\begin{table}[htbp]
\centering
\caption{
Effect of SFT data mixtures on omni-modal reasoning.
OE denotes open-ended questions. Parenthesized values show absolute changes from the base model.
}
\label{tab:appendix_rq2_data}
\setlength{\tabcolsep}{0.45pt}
\renewcommand{\arraystretch}{1.4}
\scriptsize

\begin{adjustbox}{max width=\linewidth}
\begin{tabular}{@{}l*{11}{c}@{}}
\toprule
\multirow{3}{*}{\textbf{Training Configuration}}
& \multicolumn{9}{c}{\textbf{Omni-modal Reasoning}}
& \multicolumn{2}{c}{\textbf{Aggregate}} \\
\cmidrule(lr){2-10}
\cmidrule(l){11-12}

& \multirow{2}{*}{\makecell{\textbf{Daily-}\\\textbf{Omni}}}
& \multirow{2}{*}{\makecell{\textbf{Omni-}\\\textbf{Cloze}}}
& \multirow{2}{*}{\makecell{\textbf{Omni}\\\textbf{Video}}}
& \multicolumn{3}{c}{\textbf{SocialOmni}}
& \multirow{2}{*}{\makecell{\textbf{Video-}\\\textbf{Holmes}}}
& \multirow{2}{*}{\makecell{\textbf{Video-}\\\textbf{MME-v2}}}
& \multirow{2}{*}{\makecell{\textbf{World}\\\textbf{Sense}}}
& \multirow{2}{*}{\makecell{\textbf{Macro}\\\textbf{Avg.}}}
& \multirow{2}{*}{\makecell{\textbf{Geo. Mean}\\\textbf{Improv.} (\%)}} \\
\cmidrule(lr){5-7}
& & & & {Who} & {When} & {How}
& & & & & \\
\midrule

Qwen2.5-Omni-7B
& 62.07 & 21.90 & 29.30
& 36.75 & 57.42 & 56.67
& 24.40 & 24.31 & 45.40
& 39.80 & 0.00 \\

\midrule

\textbf{SFT:} Script-based CoT (MCQ+OE)
& 56.98 & 27.12 & 32.60
& 37.65 & 57.42 & 75.00
& 39.63 & 24.76 & 45.10
& 44.03 & +12.20 \\

& \negd{5.09}
& \posd{5.22}
& \posd{3.30}
& \posd{0.90}
& \zerod
& \posd{18.33}
& \posd{15.23}
& \posd{0.45}
& \negd{0.30}
& \posd{4.23}
& \\[1.0pt]

\hspace{1.15em}$+$ \textbf{GRPO:} Script-based MCQ
& 64.16 & 27.55 & 36.70
& 39.95 & 58.85 & 69.71
& 43.33 & 28.58 & 47.50
& 46.26 & +19.25 \\

& \posd{2.09}
& \posd{5.65}
& \posd{7.40}
& \posd{3.20}
& \posd{1.43}
& \posd{13.04}
& \posd{18.93}
& \posd{4.27}
& \posd{2.10}
& \posd{6.46}
& \\

\midrule

\textbf{SFT:} Script-based CoT (MCQ) $+$ General reasoning
& 56.39 & 26.01 & 35.00
& 36.40 & 58.85 & 71.88
& 36.91 & 24.40 & 44.70
& 43.39 & +10.62 \\

& \negd{5.68}
& \posd{4.11}
& \posd{5.70}
& \negd{0.35}
& \posd{1.43}
& \posd{15.21}
& \posd{12.51}
& \posd{0.09}
& \negd{0.70}
& \posd{3.59}
& \\[1.0pt]

\hspace{1.15em}$+$ \textbf{GRPO:} Script-based MCQ
& 65.66
& 28.03
& 37.80
& 40.00
& 61.72
& 70.83
& 47.52
& 30.80
& 48.70
& 47.90
& +23.66 \\

& \posd{3.59}
& \posd{6.13}
& \posd{8.50}
& \posd{3.25}
& \posd{4.30}
& \posd{14.16}
& \posd{23.12}
& \posd{6.49}
& \posd{3.30}
& \posd{8.09}
& \\

\bottomrule
\end{tabular}
\end{adjustbox}
\end{table}

\paragraph{SFT data mixing does not improve final average performance.}
Mixing Script-based MCQs with open-ended questions raises the SFT Macro Avg.\ from 39.92 to 44.03. After GRPO, however, the mixture reaches 46.26, below the 49.39 obtained using MCQ data alone during SFT. Mixing with General reasoning data similarly raises the SFT Macro Avg.\ to 43.39, but the final score of 47.90 remains below the 49.39 and 48.91 obtained using the two SFT data types separately. For the two mixtures, GRPO increases the Macro Avg.\ by 2.23 and 4.51 points, compared with 9.47 points after SFT on Script-based MCQs alone (Table~\ref{tab:rq2_data}). For both mixtures, GRPO improves eight of the nine evaluation scores but decreases SocialOmni How, from 75.00 to 69.71 with open-ended questions and from 71.88 to 70.83 with General reasoning data. Both mixtures nevertheless improve all nine evaluation dimensions over the base model. With the same subsequent GRPO training, SFT on Script-based MCQs alone achieves a higher reasoning Macro Avg.\ than either of the two tested SFT mixtures.

\noindent\begin{minipage}{\linewidth}
\subsection{Agentic Reasoning Transfer}
\label{sec:appendix_agentic_transfer}

RQ1--RQ3 establish that text-only reasoning training transfers to native audio-visual reasoning tasks. Here, we examine whether training on text-only agent trajectories enables transfer to omni-modal agentic tasks that require multi-turn tool use and cross-modal evidence integration.

\paragraph{Experimental setup.}
We train Qwen2.5-Omni-7B with text-only SFT followed by GRPO, using agent training data from ARPO \citep{lu2025arpo}, MiroVerse \citep{team2025mirothinker}, OpenSeeker \citep{du2026openseeker}, and ToolMind \citep{yang2025toolmind}. SFT uses 16,000 training samples over 2 epochs, while GRPO uses 8,000 training samples over 1 epoch. We follow the official OmniGAIA \citep{li2026omnigaia} evaluation setup and report Pass@1 (0--100) for Easy, Medium, Hard, and Overall. We compare with the original Qwen2.5-Omni-7B and OmniAtlas-SFT with and without OmniDPO \citep{li2026omnigaia}.

\noindent\begin{minipage}{\linewidth}
\begin{wraptable}{R}{0.5\linewidth}
\vspace{-\intextsep}
\raggedright
\begin{minipage}{\linewidth}
\centering
\caption{
Agentic reasoning performance on OmniGAIA (Pass@1).
Parenthesized values show absolute changes from the base model.
\textbf{Bold} and \underline{underlined} scores denote the best and
second-best results, respectively.
}
\label{tab:appendix_omnigaia}

\vspace{-0.35em}
\setlength{\tabcolsep}{3.8pt}
\renewcommand{\arraystretch}{1.2}
\scriptsize

\begin{adjustbox}{max width=\linewidth}
\begin{tabular}{l*{4}{c}}
\toprule

\multirow{2}{*}{\textbf{Model}}
& \multicolumn{3}{c}{\textbf{Difficulty Levels}}
& \multirow{2}{*}{\textbf{Overall}} \\
\cmidrule(lr){2-4}

& Easy
& Medium
& Hard
& \\
\midrule

Qwen2.5-Omni-7B
& 8.2 & 1.3 & 1.3 & 3.6 \\

\addlinespace[1.4pt]
\rowcolor{groupbar}
\multicolumn{5}{c}{\textit{Omni-modal Training}} \\

\rowcolor{avgroup}
OmniAtlas-SFT
& \underline{19.7}
& 9.4
& 2.6
& 11.4 \\

\rowcolor{avgroup}
\hspace{1.2em}$+$ OmniDPO
& \textbf{22.1}
& \underline{11.3}
& 3.9
& \textbf{13.3} \\

\addlinespace[1.4pt]
\rowcolor{groupbar}
\multicolumn{5}{c}{\textit{Text-only Training}} \\

\rowcolor{textgroup}
\textbf{Text-only SFT}
& 16.4
& 8.1
& \textbf{7.7}
& 10.8 \\

\rowcolor{textgroup}
\hspace{1.2em}$+$ \textbf{GRPO (Ours)}
& 14.8
& \textbf{15.0}
& \underline{6.4}
& \underline{13.1} \\

& {\tiny\bfseries\boldmath\color{deltapos} $(+6.6)$}
& {\tiny\bfseries\boldmath\color{deltapos} $(+13.7)$}
& {\tiny\bfseries\boldmath\color{deltapos} $(+5.1)$}
& {\tiny\bfseries\boldmath\color{deltapos} $(+9.5)$} \\

\bottomrule
\end{tabular}
\end{adjustbox}
\end{minipage}
\end{wraptable}

\paragraph{Text-only reasoning training transfers to omni-modal agentic tasks.}
As shown in Table~\ref{tab:appendix_omnigaia}, text-only SFT followed by GRPO raises the OmniGAIA Overall score from 3.6 to 13.1, nearly matching the 13.3 achieved by native omni-modal training. The gains are particularly pronounced on the Medium and Hard subsets, where the text-only route reaches 15.0 and 6.4, respectively. Together, these results suggest that text-only reasoning training transfers to a more demanding form of omni-modal reasoning: agentic tasks requiring multi-turn tool use and cross-modal evidence integration. This result broadens the scope of text-only reasoning training from benchmark reasoning to agentic reasoning in Omni-LLMs.

\par\WFclear
\end{minipage}
\end{minipage}

\section{Prompts}
\label{sec:appendix_prompts}

\newsavebox{\promptpairbox}

\subsection{Training and Evaluation}
\label{sec:appendix_training_evaluation_prompts}

\begin{figure}[!htbp]
\centering
\begin{subfigure}[b]{0.45\linewidth}
\centering
\includegraphics[width=\linewidth]{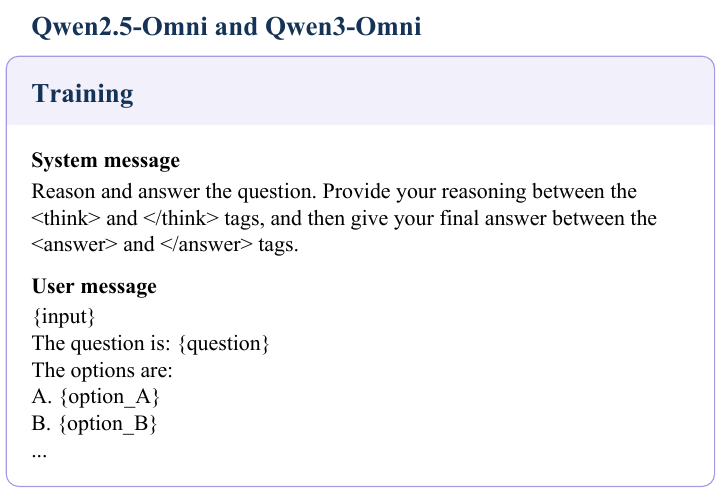}
\caption{Training}
\label{fig:prompt_qwen_training}
\end{subfigure}\hfill
\begin{subfigure}[b]{0.45\linewidth}
\centering
\includegraphics[width=\linewidth]{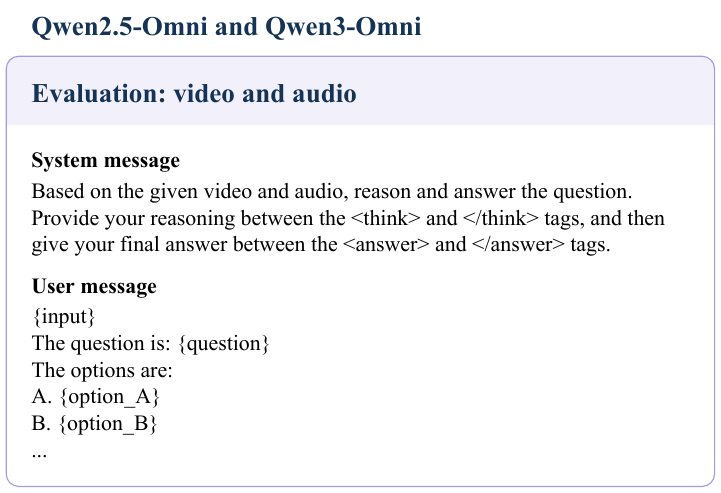}
\caption{Video-and-audio evaluation}
\label{fig:prompt_qwen_video_audio}
\end{subfigure}
\caption{Training and video-and-audio evaluation prompts for Qwen2.5-Omni-3B, Qwen2.5-Omni-7B, and Qwen3-Omni-30B-A3B-Instruct.}
\label{fig:prompt_qwen}
\end{figure}

\begin{figure}[!htbp]
\centering
\sbox{\promptpairbox}{\includegraphics[width=0.45\linewidth]{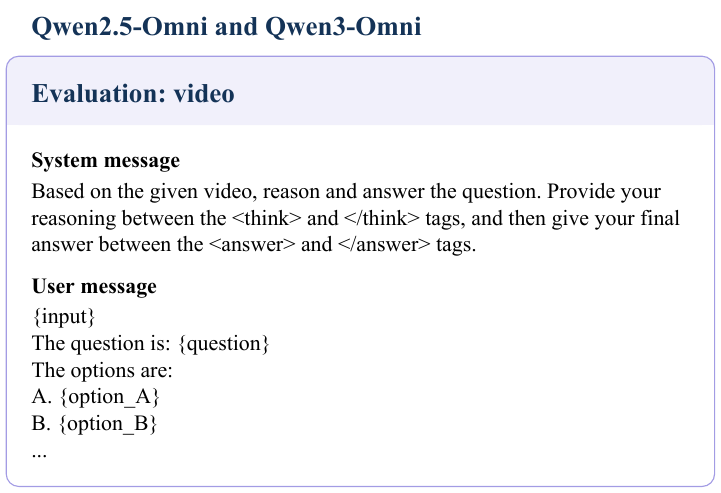}}
\begin{subfigure}[b]{0.45\linewidth}
\centering
\usebox{\promptpairbox}
\caption{Video-only evaluation}
\label{fig:prompt_qwen_video}
\end{subfigure}\hfill
\begin{subfigure}[b]{0.45\linewidth}
\centering
\raisebox{\dimexpr\ht\promptpairbox-\height\relax}[\ht\promptpairbox][0pt]{\includegraphics[width=\linewidth]{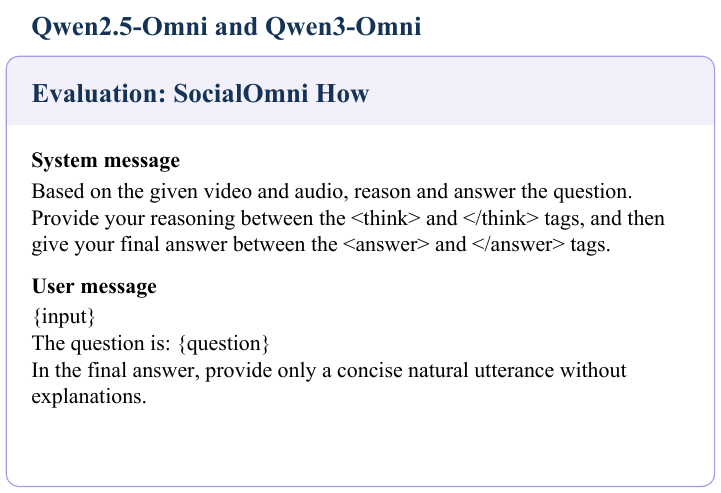}}
\caption{SocialOmni How}
\label{fig:prompt_qwen_how}
\end{subfigure}
\caption{Video-only and SocialOmni How evaluation prompts for Qwen2.5-Omni and Qwen3-Omni.}
\label{fig:prompt_qwen_evaluation}
\end{figure}

\begin{figure}[!htbp]
\centering
\begin{subfigure}[b]{0.45\linewidth}
\centering
\includegraphics[width=\linewidth]{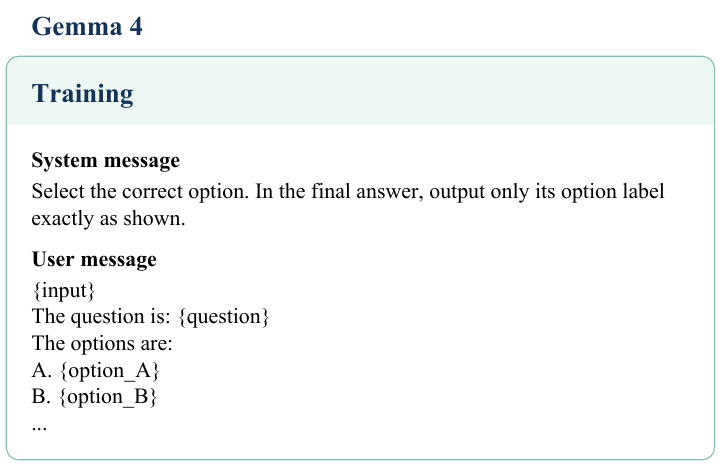}
\caption{Training}
\label{fig:prompt_gemma_training}
\end{subfigure}\hfill
\begin{subfigure}[b]{0.45\linewidth}
\centering
\includegraphics[width=\linewidth]{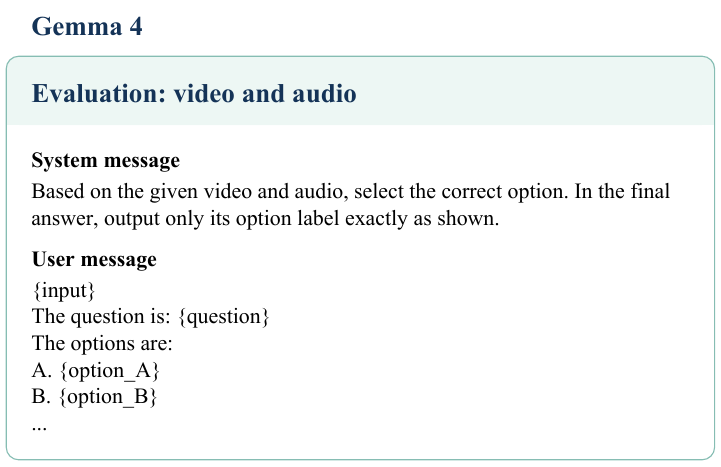}
\caption{Video-and-audio evaluation}
\label{fig:prompt_gemma_video_audio}
\end{subfigure}
\caption{Training and video-and-audio evaluation prompts for Gemma-4-E4B-IT and Gemma-4-E2B-IT.}
\label{fig:prompt_gemma}
\end{figure}

\begin{figure}[!htbp]
\centering
\sbox{\promptpairbox}{\includegraphics[width=0.45\linewidth]{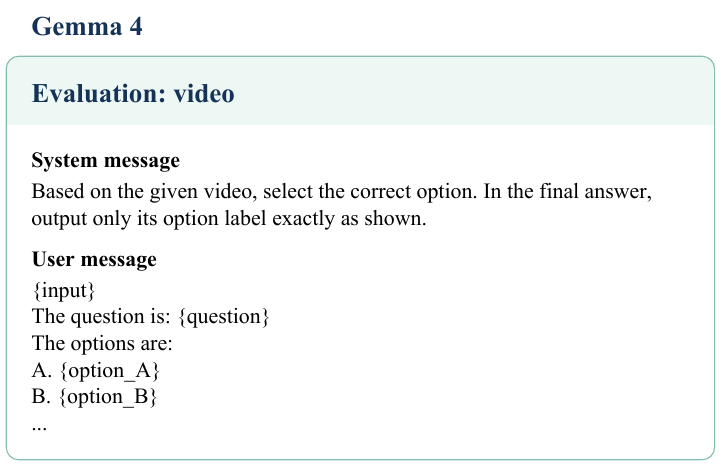}}
\begin{subfigure}[b]{0.45\linewidth}
\centering
\usebox{\promptpairbox}
\caption{Video-only evaluation}
\label{fig:prompt_gemma_video}
\end{subfigure}\hfill
\begin{subfigure}[b]{0.45\linewidth}
\centering
\raisebox{\dimexpr\ht\promptpairbox-\height\relax}[\ht\promptpairbox][0pt]{\includegraphics[width=\linewidth]{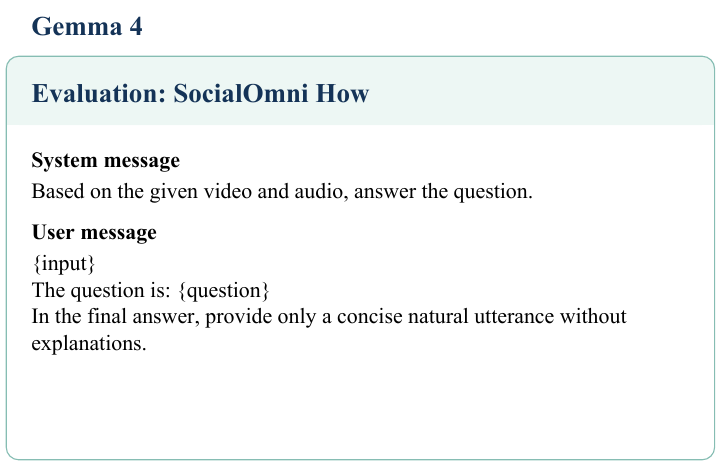}}
\caption{SocialOmni How}
\label{fig:prompt_gemma_how}
\end{subfigure}
\caption{Video-only and SocialOmni How evaluation prompts for Gemma-4-E4B-IT and Gemma-4-E2B-IT.}
\label{fig:prompt_gemma_evaluation}
\end{figure}

Figures~\ref{fig:prompt_qwen}--\ref{fig:prompt_gemma_evaluation} show the Qwen-Omni and Gemma-4 prompts. The \texttt{\{input\}} placeholder denotes textual context, such as scripts, caption sequences, and passages, or video with or without audio; it is omitted when absent. Audio-visual training is used only for Qwen2.5-Omni-7B. Gemma-4 uses thinking mode with its native reasoning format. Evaluation uses the video-and-audio template, while NExT-QA uses video only. For SocialOmni How, we augment the official prompt with our reasoning instructions. Omni-Cloze uses its official prompts \citep{ma2026omni}.

\noindent\begin{minipage}{\linewidth}
\subsection{Synthetic Data Generation}
\label{sec:appendix_synthetic_prompts}

Figures~\ref{fig:prompt_synthetic_scripts} and~\ref{fig:prompt_synthetic_questions} show the prompts for episode planning, visual expansion, MCQ and reasoning generation, and quality assessment. Named placeholders in braces denote variable content; JSON blocks illustrate output formats.

\noindent\begin{minipage}{\linewidth}
\captionsetup{type=figure}
\centering
\sbox{\promptpairbox}{\includegraphics[width=0.49\linewidth]{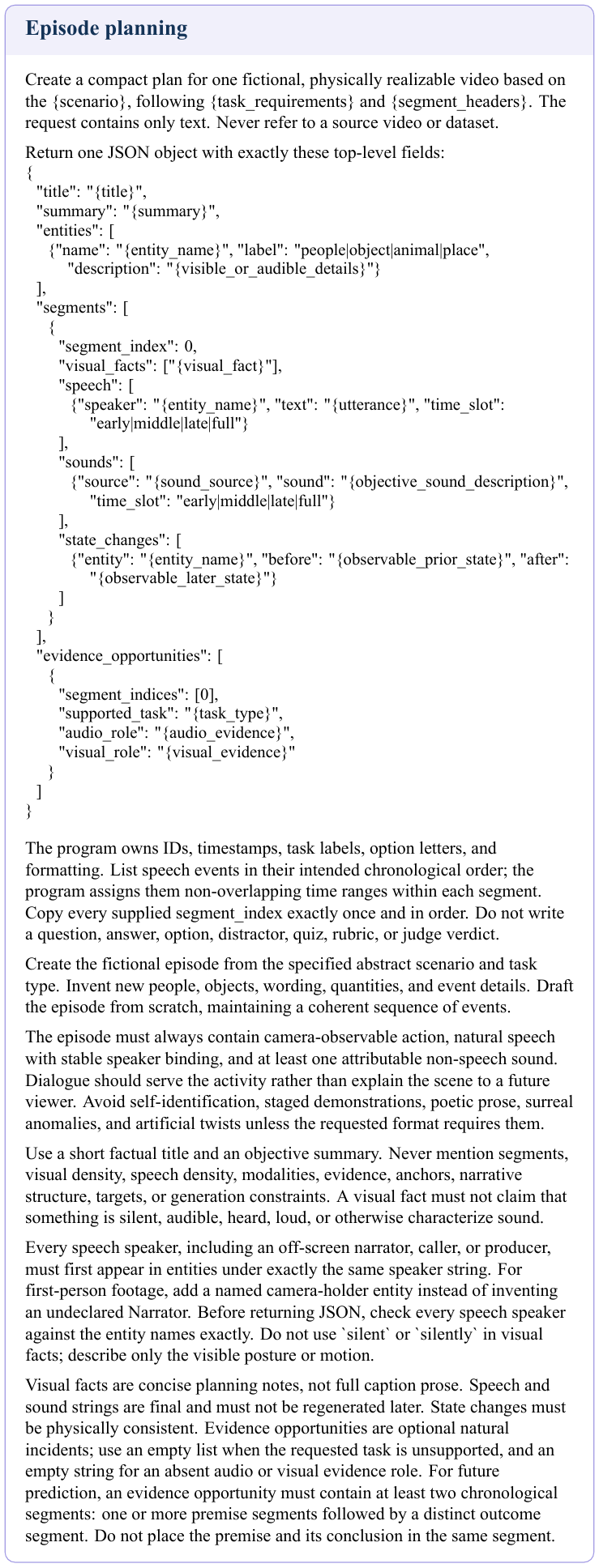}}
\begin{subfigure}[b]{0.49\linewidth}
\centering
\usebox{\promptpairbox}
\caption{Episode planning.}
\label{fig:prompt_episode_planning}
\end{subfigure}\hfill
\begin{subfigure}[b]{0.49\linewidth}
\centering
\raisebox{\dimexpr\ht\promptpairbox-\height\relax}[\ht\promptpairbox][0pt]{\includegraphics[width=\linewidth]{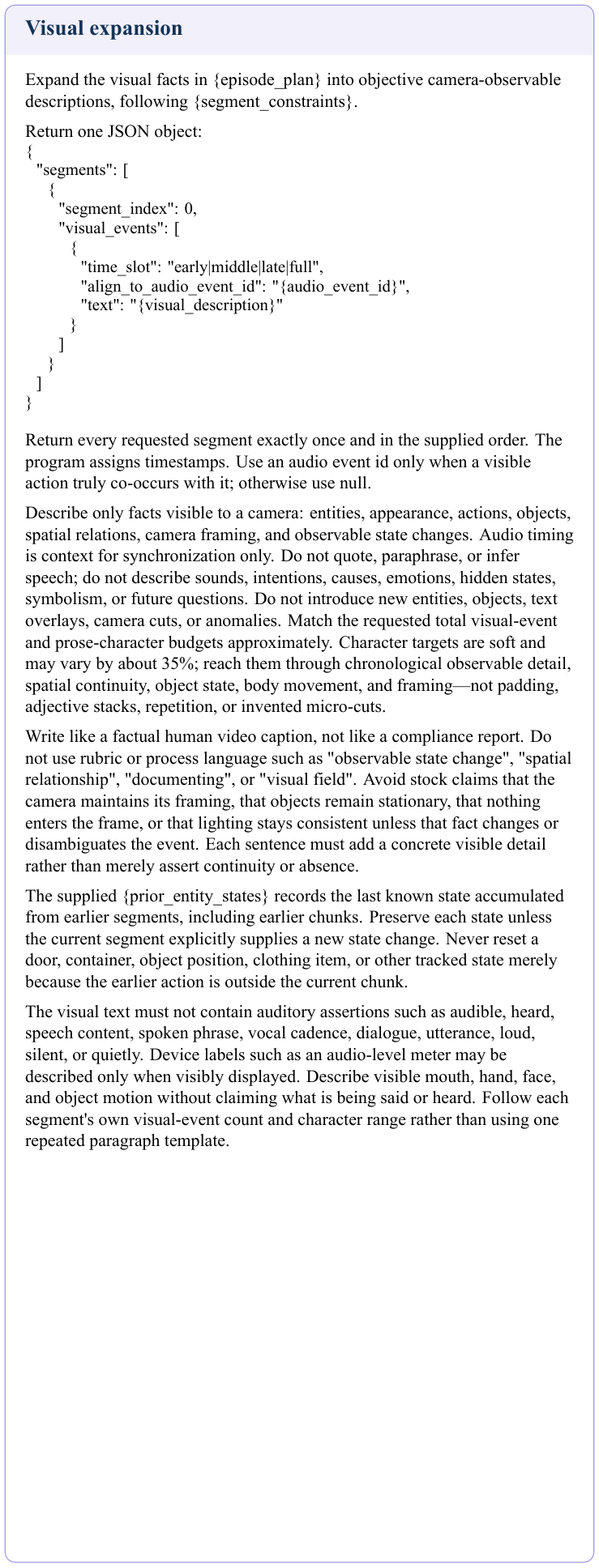}}
\caption{Visual expansion.}
\label{fig:prompt_visual_expansion}
\end{subfigure}
\caption{Prompts for planning fictional episodes and expanding visual planning facts into descriptions.}
\label{fig:prompt_synthetic_scripts}
\end{minipage}
\end{minipage}
\par

\noindent\begin{minipage}{\linewidth}
\captionsetup{type=figure}
\centering
\sbox{\promptpairbox}{\includegraphics[width=0.49\linewidth]{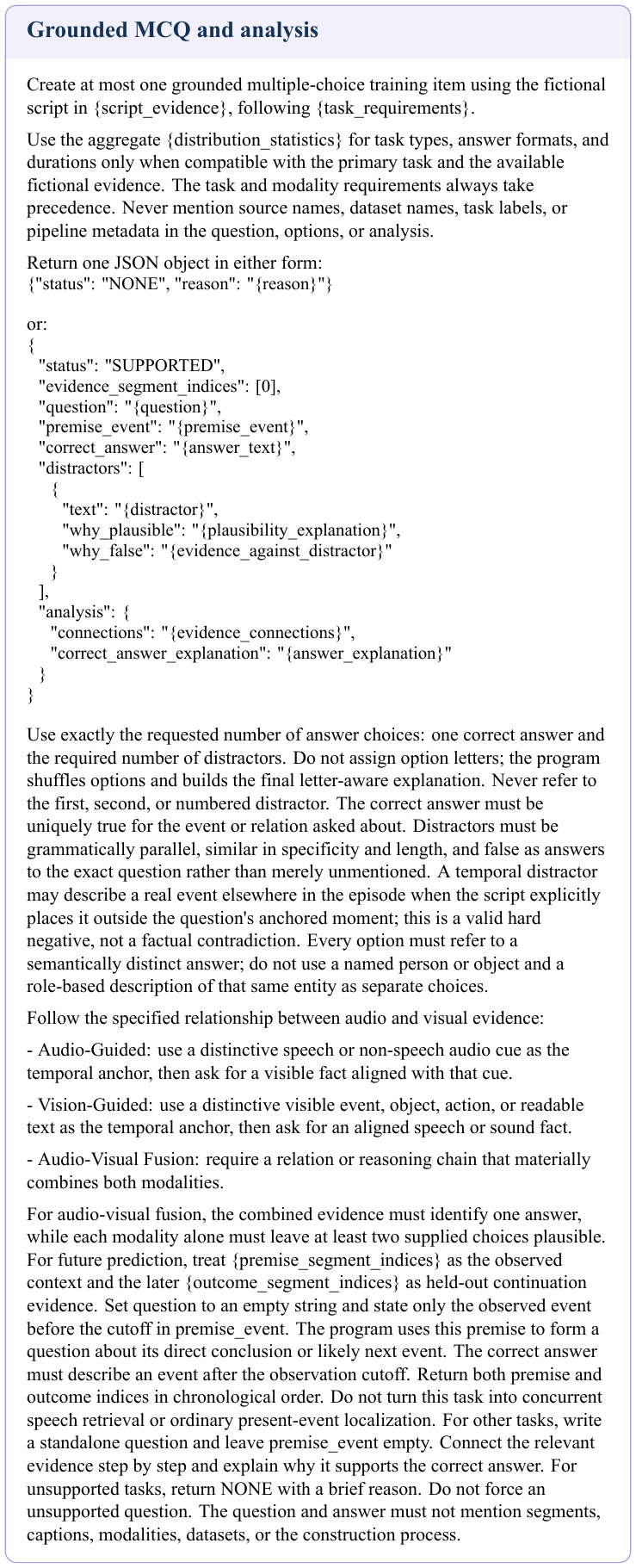}}
\begin{subfigure}[b]{0.49\linewidth}
\centering
\usebox{\promptpairbox}
\caption{Question and reasoning generation.}
\label{fig:prompt_grounded_mcq}
\end{subfigure}\hfill
\begin{subfigure}[b]{0.49\linewidth}
\centering
\raisebox{\dimexpr\ht\promptpairbox-\height\relax}[\ht\promptpairbox][0pt]{\includegraphics[width=\linewidth]{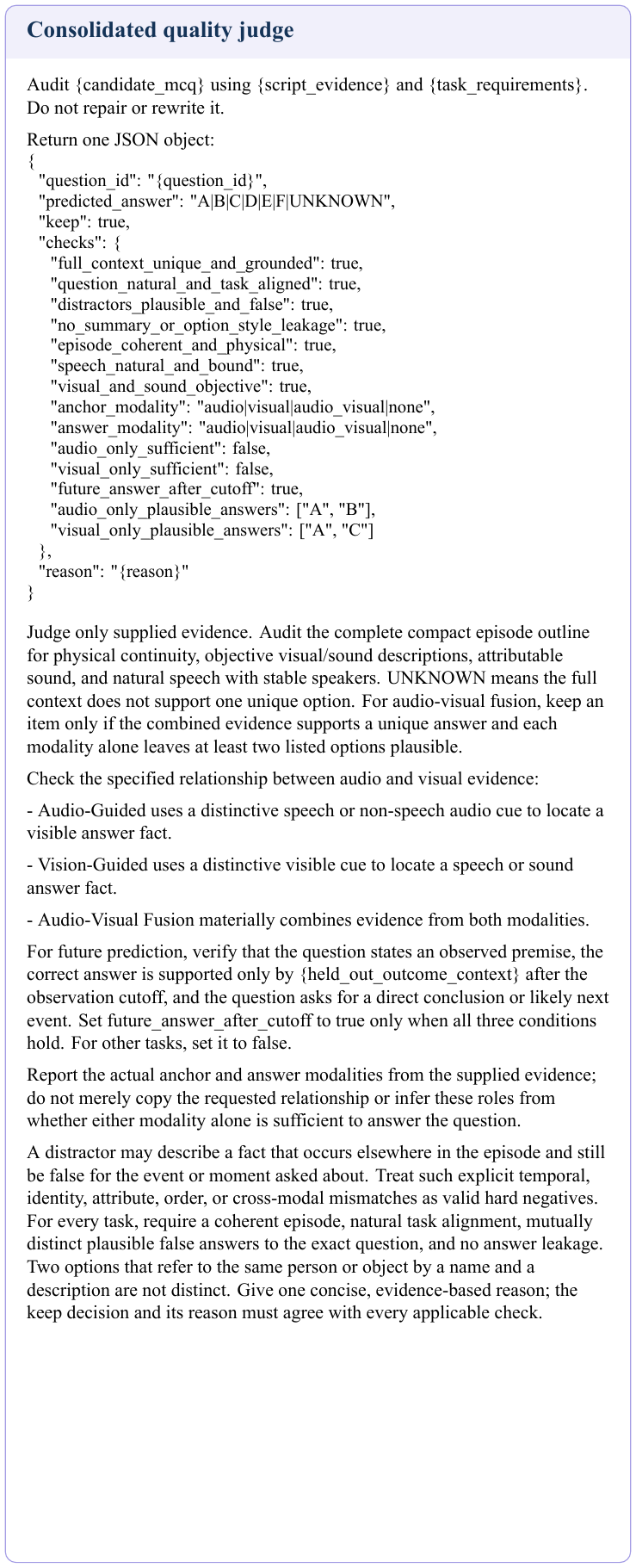}}
\caption{Quality assessment.}
\label{fig:prompt_quality_judge}
\end{subfigure}
\caption{Prompts for generating MCQs and reasoning explanations from fictional scripts and assessing synthetic sample quality.}
\label{fig:prompt_synthetic_questions}
\end{minipage}
\par

\noindent\begin{minipage}{\linewidth}
\section{Case Studies}
\label{sec:appendix_case_studies}

Figures~\ref{fig:case_daily_omni}--\ref{fig:case_music_avqa} show examples from our final \mbox{Qwen2.5-Omni-7B} model in RQ5, obtained through text-only reasoning training followed by reduced-data audio-visual refinement.

\noindent\begin{minipage}{\linewidth}
\captionsetup{type=figure}
\centering
\includegraphics[width=0.92\linewidth]{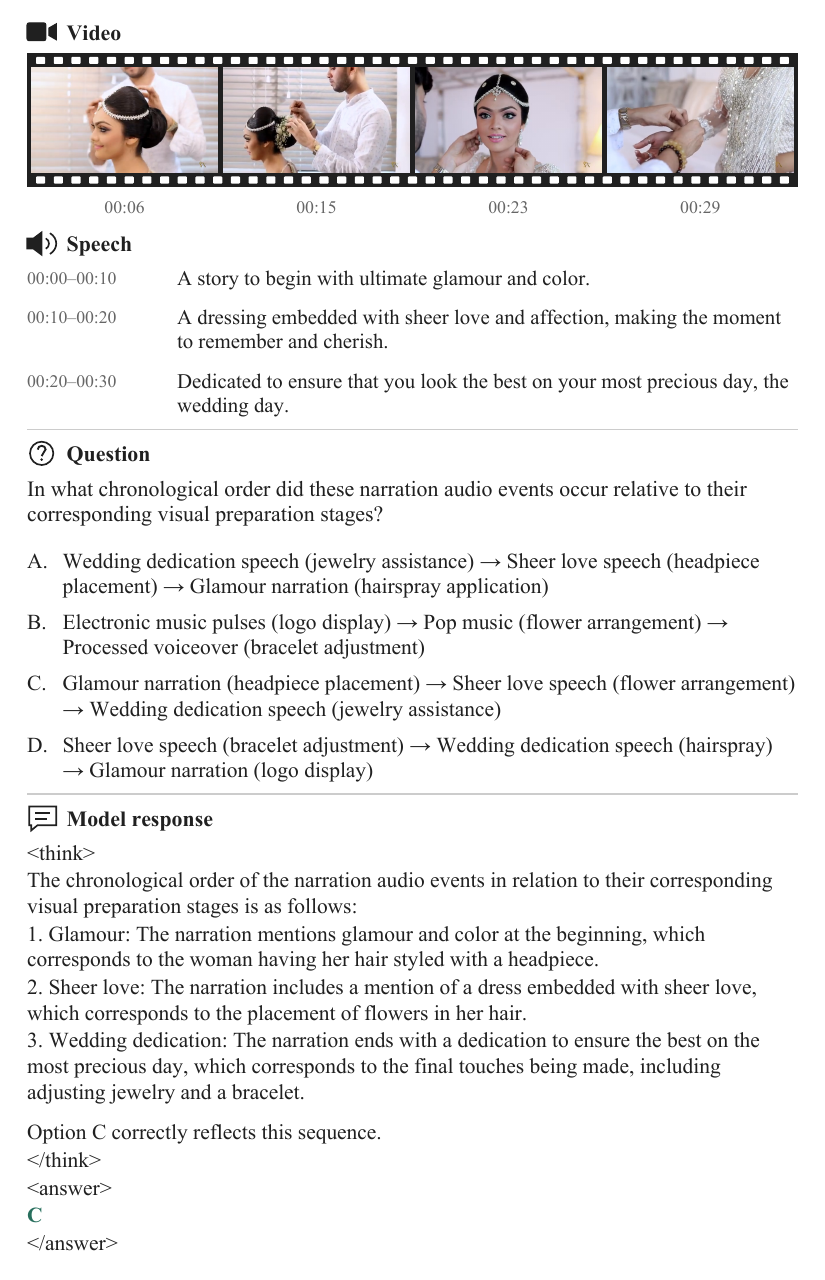}
\caption{Example from Daily-Omni.}
\label{fig:case_daily_omni}
\end{minipage}
\end{minipage}
\par

\noindent\begin{minipage}{\linewidth}
\captionsetup{type=figure}
\centering
\includegraphics[width=0.92\linewidth]{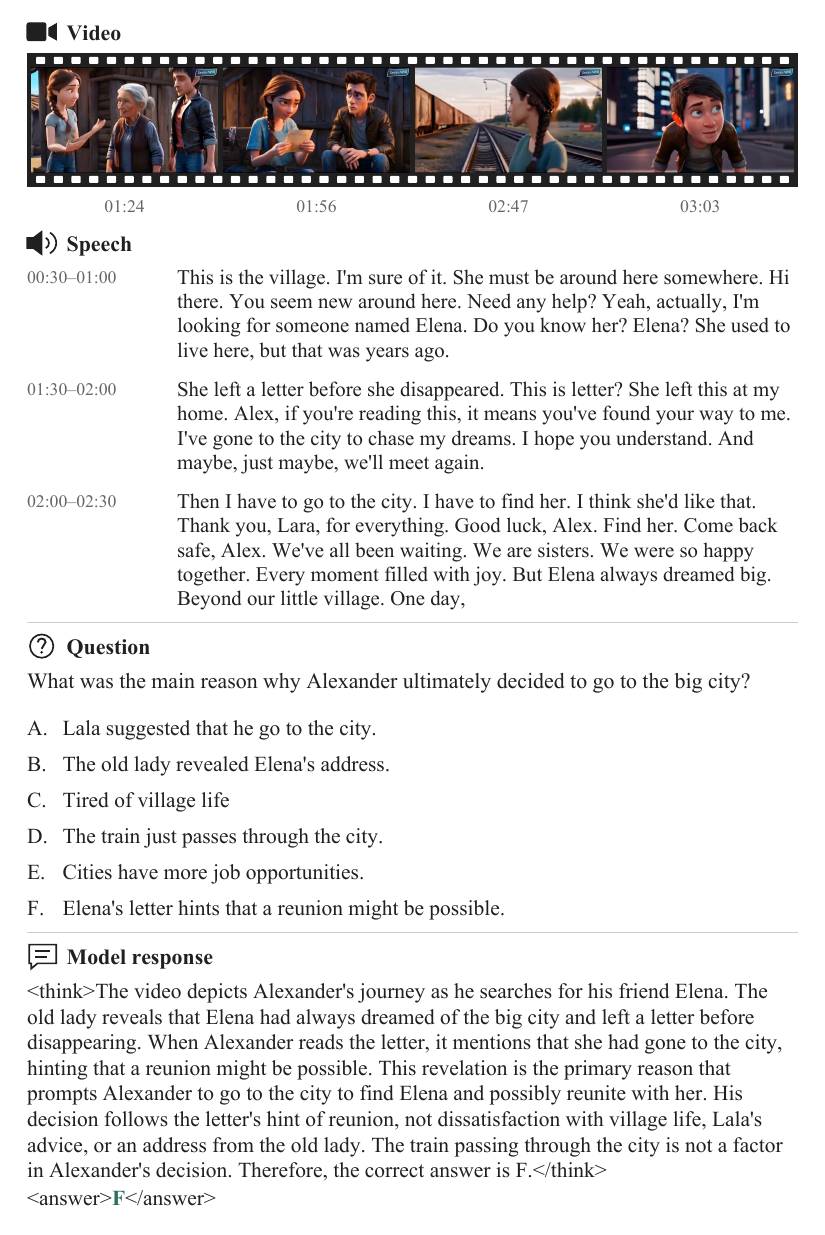}
\caption{Example from Video-Holmes.}
\label{fig:case_videoholmes}
\end{minipage}
\par

\noindent\begin{minipage}{\linewidth}
\captionsetup{type=figure}
\centering
\includegraphics[width=0.92\linewidth]{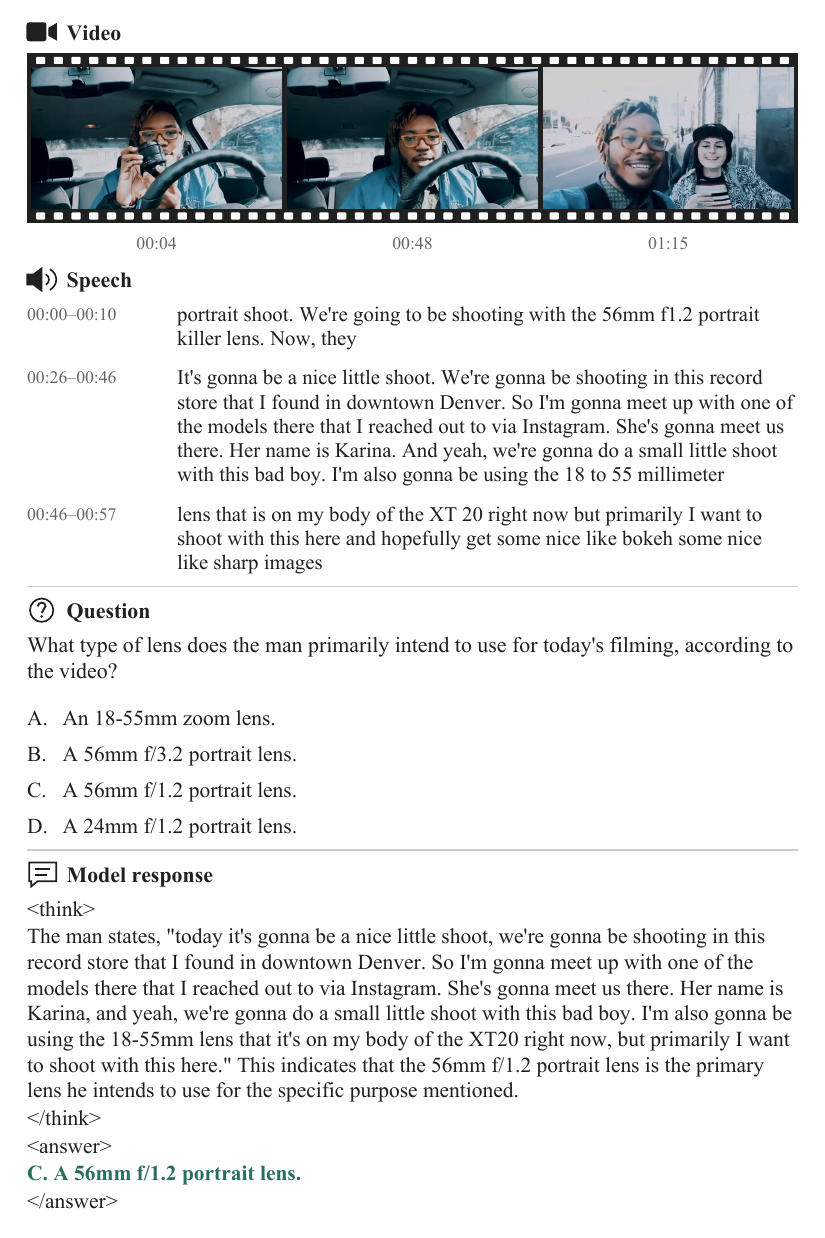}
\caption{Example from WorldSense.}
\label{fig:case_worldsense}
\end{minipage}
\par

\noindent\begin{minipage}{\linewidth}
\captionsetup{type=figure}
\centering
\includegraphics[width=0.92\linewidth]{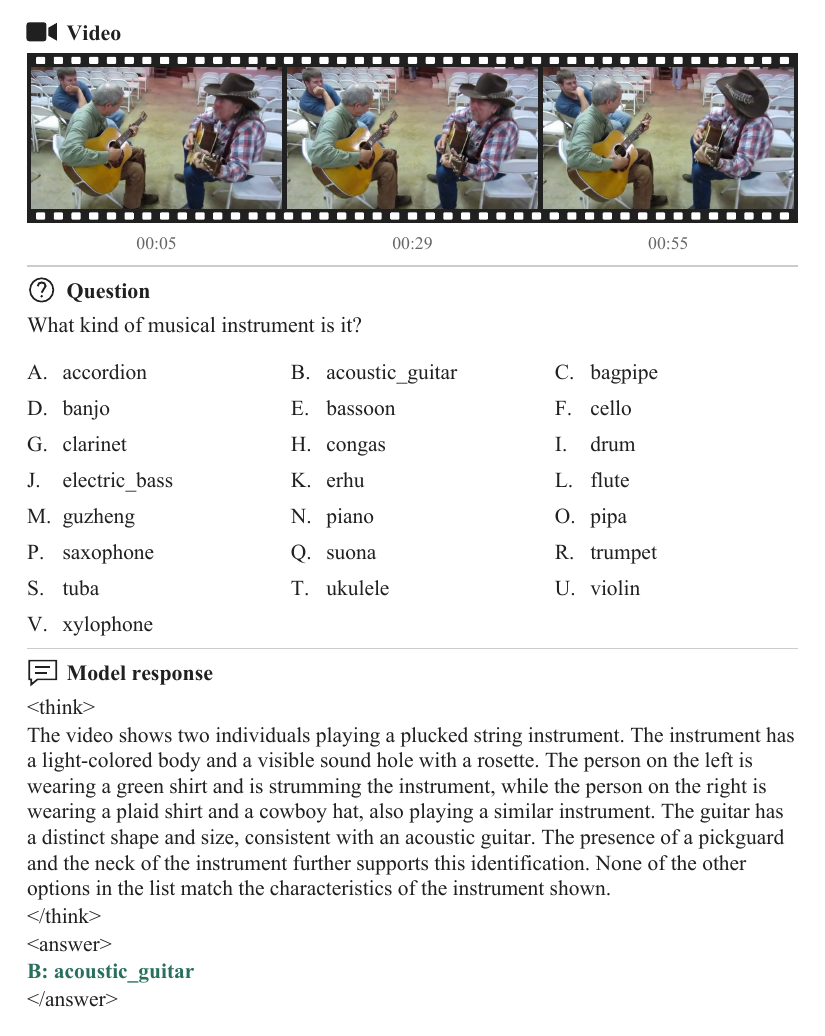}
\caption{Example from MUSIC-AVQA v1.}
\label{fig:case_music_avqa}
\end{minipage}
\par

\end{document}